\documentclass{article}

\usepackage{arxiv}

\usepackage[utf8]{inputenc} 
\usepackage[T1]{fontenc}    
\usepackage{hyperref}       
\usepackage{url}            
\usepackage{booktabs}       
\usepackage{amsfonts}       
\usepackage{amsmath} 
\usepackage{amssymb}  
\usepackage{nicefrac}       
\usepackage{microtype}      
\usepackage{cleveref}       
\usepackage{lipsum}         
\usepackage{graphicx}
\usepackage{natbib}
\usepackage{doi}
\usepackage{epsfig} 
\usepackage{mathptmx} 
\usepackage{times} 
\usepackage{bm}

\usepackage{subfig}
\usepackage{multirow}
\usepackage[export]{adjustbox}
\usepackage{url}
\usepackage[usenames]{color}
\usepackage[mathscr]{euscript}

\title{Atlanta Scaled Layouts from Non-central Panoramas}

\date{}

\author{ \href{https://orcid.org/0000-0003-2674-4844}{\includegraphics[scale=0.06]{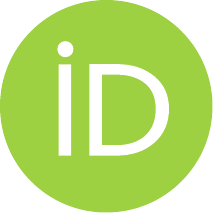}\hspace{1mm}Bruno Berenguel-Baeta}\thanks{Corresponding author.} \\
	Instituto de Investigacion en Ingenieria de Aragon\\
	Department of Computer Science and Systems Engineering\\
	University of Zaragoza\\
	Zaragoza, Spain \\
	\texttt{berenguel@unizar.es} \\
	\And
	\href{https://orcid.org/0000-0002-8479-1748}{\includegraphics[scale=0.06]{orcid.pdf}\hspace{1mm}Jesus Bermudez-Cameo} \\
	Instituto de Investigacion en Ingenieria de Aragon\\
	Department of Computer Science and Systems Engineering\\
	University of Zaragoza\\
	Zaragoza, Spain \\
	\texttt{bermudez@unizar.es} \\
	\And
	\href{https://orcid.org/0000-0001-5209-2267}{\includegraphics[scale=0.06]{orcid.pdf}\hspace{1mm}Jose J. Guerrero} \\
	Instituto de Investigacion en Ingenieria de Aragon\\
	Department of Computer Science and Systems Engineering\\
	University of Zaragoza\\
	Zaragoza, Spain \\
	\texttt{josechu.guerrero@unizar.es} \\
}
\renewcommand{\shorttitle}{\textit{Scaled Layouts}}
\newcommand\blfootnote[1]{%
  \begingroup
  \renewcommand\thefootnote{}\footnote{#1}%
  \addtocounter{footnote}{-1}%
  \endgroup
}
\hypersetup{
pdftitle={Atlanta Scaled Layouts from Non-central Panoramas},
pdfsubject={cs.CV},
pdfauthor={Bruno Berenguel-Baeta, Jesus Bermudez-Cameo, Jose J. Guerrero},
pdfkeywords={Omnidirectional Vision; 3D Vision; Non-central Cameras; Layout recovery; Scene understanding},
}

\begin{document}
\maketitle
\blfootnote{A final version of this article can be found at \url{https://doi.org/10.1016/j.patcog.2022.108740}}

\begin{abstract}
	In this work we present a novel approach for 3D layout recovery of indoor environments using a non-central acquisition system. From a non-central panorama, full and scaled 3D lines can be independently recovered by geometry reasoning without geometric nor scale assumptions. However, their sensitivity to noise and complex geometric modeling has led these panoramas being little investigated.
Our new pipeline aims to extract the boundaries of the structural lines of an indoor environment with a neural network and exploit the properties of non-central projection systems in a new geometrical processing to recover an scaled 3D layout.
The results of our experiments show that we improve state-of-the-art methods for layout reconstruction and line extraction in non-central projection systems. 
We completely solve the problem in Manhattan and Atlanta environments, handling occlusions and retrieving the metric scale of the room without extra measurements.
As far as the authors knowledge goes, our approach is the first work using deep learning on non-central panoramas and recovering scaled layouts from single panoramas.

\end{abstract}


\section{Introduction}
\label{sec:intro}

Layout recovery and 3D understanding of indoor environments is a hot topic in computer vision research \cite{zou2021manhattan}. Recovering the information of an environment from a single view is an attractive tool for different applications such as virtual or augmented reality \cite{karsch2011rendering} and human pose estimation \cite{fouhey2012people}. 
First approaches for layout recovery relied on pure geometrical processing. Those methods usually required hard layout assumptions and iterative proccesses in order to obtain proper results. Besides, since many hypotheses and verifications should be made, these approaches derive in very slow implementations, not suitable for real time applications.
The development of neural networks made the problem of layout recovery more accurate, efficient and faster. The high and low level features obtained by deep learning architectures have proven to be useful for structural recovery of indoor environments.

Through the development of algorithms for layout recovery, different kinds of acquisition systems have been used in order to obtain more information of the environment with as less images as possible. An example is the evolution from perspective images to equirectangular panoramas, which acquire much more information of the environment in a single image.With this extra information, better reconstructions from more complex layouts could be made. 

\begin{figure}
	\centering
	\includegraphics[width=0.85\textwidth]{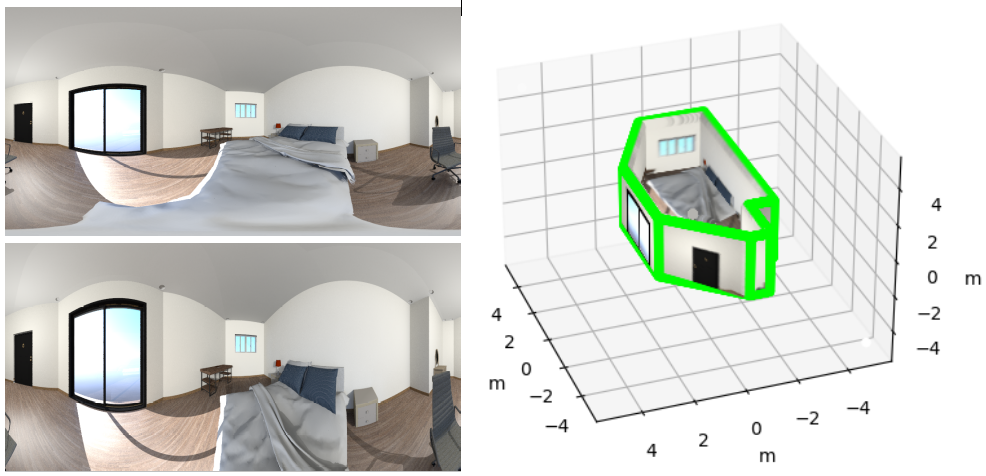}
	\caption{Central (top-left) and non-central (bottom-left) panoramas from the same virtual environment taken in the same position. Both panoramas have similar appearance but there are subtle differences in favor of the second if we want to obtain 3D information.
	On the right, the scaled layout from a single non-central panorama in Atlanta world.
	The green wireframe shows the real 3D layout of the virtual environment.}
	\label{fig:intro}
\end{figure}

In this paper, we propose to go a step further and evolve taking as acquisition system the non-central panoramas proposed in \cite{li2004stereo, menem2004constraints}. These panoramas provide 360 information of the environment and the image distortion of the non-central acquisition systems includes subtle differences allowing geometric 3D reasoning. In particular, the distortion of the curves fitting the projections of lines encodes the full 3D description of the line. This characteristic of non-central projection systems is a clear advantage for environment reconstruction, since it allows to recover the scale of the environment directly from the image, without measurement assumptions (e.g. camera position or room height). However, due to their sensitivity to noise and complex geometric reasoning, non-central panoramas have been little investigated.
Figure \ref{fig:intro} shows two panoramas, one central and the other non-central, in the same environment and from the same position. From the non-central panorama we can recover a more accurate layout, including the real scale.

In this work we present the first proposal of layout recovery with single non-central panoramas and the first deep learning approach for this kind of images. 
We propose to adapt the neural network architecture of HorizonNet \cite{sun2019horizonnet} to non-central circular panoramas for the extraction of the boundaries of the structural lines from indoor environments. As in \cite{sun2019horizonnet}, we assume that our panoramas are horizontally oriented and that the layout share the ceiling and floor heights. These restrictions will give strong priors in the geometrical proccessing.
Taking advantage of the omnidirectional view of non-central panoramas and the unique properties of the non-central projection systems, we extract the 3D information of the structural lines provided by the network. The experiments performed show that our pipeline outperforms the state of the art in layout reconstruction by a margin.
The main contributions of this paper are as follow. 

\begin{itemize}
\item Two new geometrical solvers to obtain the layout of an environment in a Manhattan or an Atlanta world assumption for non-central projection systems. 
\item First work that use deep learning with non-central projection systems.
\item First work that extracts the layout scale from a single panorama without extra measurements.
\item Scaled layout recovery from Manhattan and Atlanta world assumptions, handling occlusions, from a single non-central panorama.
\item We also present the first data-set of non-central panoramas in the state of the art.
\end{itemize}

The next section presents an overview of the state of the art in layout reconstruction and non-central projection systems. In Section \ref{sec:noncentralmodel}, we introduce the non-central circular panorama. 
Section \ref{sec:proposal} present the pipeline of our proposal and details the neural network architecture used.
The geometrical proccessing and the full pipeline are detailed in Section \ref{sec:geometry}. 
Finally, we present a set of experiments to validate our proposal and we also make a comparison with state-of-the-art methods in Section \ref{sec:experimets}. We also present a data-set of non-central circular panoramas used for train our network and perform the experiments. 
Preliminary results of our work have been presented in \cite{berenguel2021scaled}. Here, the proposed approach is revised and completed, including an ablation study, a more complete experimentation and comparison with state of the art methods and more detailed explanations of the basis and implementation.

\section{State of the art}
\label{sec:sota}


Non-central projection systems have been extensely studied from their different acquisition systems. 
Works as \cite{agrawal2010analytical, agrawal2011beyond, lopez2010catadioptric, bermudez2014line} have set the fundamentals on catadioptric systems based on conical or spherical mirrors. 
Other non-central images come from moving cameras, as the pushbroom camera \cite{gupta1997linear} or the non-central circular panorama \cite{li2004stereo, menem2004constraints}. These non-central projection systems present geometrical properties that allow to recover 3D information from single images with geometric reasoning. 
In particular, several works exploit that a 3D line can be recovered with scale from a single non-central projection \cite{gasparini2011line}.
The fundamentals for this approach consist in computing the intersection of a line by four generic rays \cite{teller1999determining}. 
However, although it is theoretically possible to recover the scaled, full 3D reconstruction of a line from a single non-central projection, in practice the results are so sensitive to noise and therefore it is not possible to directly use these approaches with current non-central systems. For this reason, more recent works aim to improve the accuracy of 3D lines fitting by imposing structural constraints. 
As example, in \cite{bermudez2016line} the line extraction is constrained to lines parallel to a known plane, which can be used for extracting horizontal lines from a non-oriented camera by using the gravity direction, for example from an IMU, as prior. Other constraints such as parallel lines or intersecting orthogonal lines are well studied and solved in \cite{bermudez2014minimal}, where they present a minimal solution to be included in a robust approach.

When the geometric constraints of structural lines are globally considered, lines and their intersections are enclosed in the concept of layout. The layout of indoor environments provides a strong prior for many computer vision tasks. Several works on virtual or augmented reality \cite{karsch2011rendering}, object recognition \cite{bao2011toward,song2016deep,nie2020shallow2deep} and human pose estimation \cite{fouhey2012people} rely on information of the environment, which is more easily obtainable once the layout is known. Many different methods have been developed in order to extract the layout of a room from different central cameras. Particularly, in recent years, the use of omnidirectional central images is on the rise, since a single image can provide enough information to make an estimation for a whole room \cite{fukano2016room, rao2021omnilayout}.
One of the first attempts for layout estimation is the work \cite{zhang2014panocontext} which presents an implementation where many 3D layout hypotheses are generated and then ranked by a Support Vector Machine (SVM). Then the best ranked hypotheses are selected and compared with the input image to test its validity.
More recent approaches take advantage of neural networks to extimate the layouts in a more efficient way. Corners for Layouts (CFL) \cite{fernandez2020corners} uses an encoder-decoder architecture with convolutions adapted to the spherical distortion of the equirectangular panorama. The output of the network are two heat-maps for the corners and edges that compose the structure of rooms. With a post-processing of this information, an up-to-scale reconstruction of Manhattan environments can be obtained.
Other recent approaches combine the convolutional networks with recurrent neural networks, which allow to obtain dependencies along the image, extracting the boundaries of the structural lines. Relying on different geometry constraints, HorizonNet \cite{sun2019horizonnet} and AtlantaNet \cite{pintore2020atlantanet} obtain a 1D representation of corners, as a probability of having a wall-wall intersection at a certain image column, and other 1D representation of the ceiling-wall and floor-wall intersections which form the structural lines of the room. This minimal representation allows to obtain a more precise approximation of the layout of the room. As in other works, after a post-processing, an up-to-scale layout reconstruction can be obtained.

In our proposal, we overcome the problem of structural lines extraction in non-central panoramas adapting the neural network of HorizonNet \cite{sun2019horizonnet}. Then, we propose a new geometrical processing, which takes advantage of two new solvers that fit Manhattan and Atlanta layouts, to recover the 3D layout of indoor environments. Besides, exploiting the geometrical properties of non-central projection systems in our geometrical processing, we are able to recover the scale of the 3D layout without extra measurements, which no state-of-the-art method is able to do.

\section{The non-central circular panorama}
\label{sec:noncentralmodel}

Central projection systems are those acquisition systems where each projecting ray that forms the image intersects through a single point, called optical center. The pinhole camera model or the spherical panorama are examples of central projection systems. By contrast, non-central projection systems do not have a unique optical center, that means, the rays that form the images do not pass through a unique optical center. This nature leads to a harder management of the information, being the result more sensitive to noise. Nevertheless, this characteristic allows to obtain more geometric information from the image than from central projection systems. In particular, we can extract the 3D information of lines directly from the image.

\begin{figure}
	\centering
	\includegraphics[width = 0.65\textwidth]{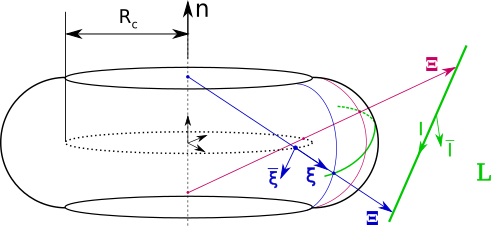}
	\caption{Toroidal projection of non-central circular panoramas. For each point of the circular trajectory of optical centers with radius $R_c$, there is a ring in which the projection is central. A 3D line $\mathbf{L}$ is projected in the toroidal surface. A projecting ray $\boldsymbol{\Xi}$ intersects the line giving a point in the toroidal surface for a unique optical center.}
	\label{fig:toro}
\end{figure}

The non-central panorama is a projection model, presented in \cite{shum1999stereo}, with symmetry of revolution in which each projecting ray intersects both an axis $\mathbf{n}$ and a circle of radius $R_c$. In the last two decades some works have studied the geometrical properties of non-central panoramas \cite{menem2004constraints, bakstein2001overview} and other non-central projection systems \cite{agrawal2010analytical,perdigoto2013calibration}. Taking advantage of the unique geometrical properties of this acquisition systems, different applications have been already studied, as 3D line extraction \cite{bermudez2016line,bermudez2017exploiting}.

The non-central circular panorama can be modeled as the projection of the environment into a toroidal surface (Fig. \ref{fig:toro}). The optical center is distributed in a circle of radius ${R_c}$, centered in an axis $\mathbf{n}$. For each optical center, we have a region where the projection is locally central, which correspond with one column in the panoramic image. We use Plücker coordinates \cite{pottmann2009computational} to define the backward projection function of the system as well as the 3D lines in the environment.

The notation to define the projection model and the math presented in this paper is as follows. For projecting rays and lines, defined in Plücker coordinates, we will use bold uppercase letters (e.g. $\boldsymbol{\Xi}$, $\mathbf{L}$). For vectors that belong to $\mathbb{R}^3$, bold lower case letters (e.g. $\mathbf{n}$), while vectors of greater dimmensions are presented as upper case letters in Euler font (e.g. $\mathscr{W}$). Matrices are presented in the serif font as uppercase letters (e.g. $\mathsf{A}$). Scalar values are presented as standard text (e.g. $R_c$, $\varphi$).

The forward projection provides the pixel coordinates $(i,j)$ for each 3D point $(x,y,z,w)^T$ defined in homogeneous coordinates. Each point is defined by two angles $(\phi,\varphi)$ from its corresponding optical center as defined in equation \eqref{eq:forward}. The angles are transformed into pixel coordinates \eqref{eq:img} taking into account the image resolution $(m_{rows},n_{columns})$ and the horizontal $(\phi_{ini},\phi_{end})$ and vertical $(\varphi_{ini},\varphi_{end})$ fields of view.

\begin{equation} \label{eq:forward}
	\varphi = atan2(y,x); \;\;\;
	\phi = atan\left(\dfrac{z}{\sqrt{x^2+y^2}-w R_c}\right)
\end{equation}

\begin{equation} \label{eq:img}
	j = n_{columns} \dfrac{\varphi - \varphi_{ini}}{\varphi_{end} - \varphi_{ini}}; \;\;\;
	i = m_{rows} \dfrac{\phi - \phi_{ini}}{\phi_{end} - \phi_{ini}}
\end{equation}

The backward projection model provides the projecting rays in Plücker coordinates from each pixel in the non-central panorama. The pixel coordinates are transformed into spherical coordinates \eqref{eq:backproj} taking into account the image resolution and field of views. Then, the projecting ray \eqref{eq:pluckerray} is computed considering the radius $R_c$ of the non-central acquisition system.

\begin{equation} \label{eq:backproj}
	\varphi = j \dfrac{\varphi_{end} - \varphi_{ini}}{n_{columns}} + \varphi_{ini}; \;\;\;
	\phi =  i \dfrac{\phi_{end} - \phi_{ini}}{m_{rows}} + \phi_{ini}
\end{equation}

\begin{equation} \label{eq:pluckerray}
	\boldsymbol{\Xi} = \begin{pmatrix}
		\boldsymbol{\xi} \\ \boldsymbol{\bar{\xi}}
		\end{pmatrix} = \begin{pmatrix}
		\cos\phi \cos\varphi \\
		\cos\phi \sin\varphi \\
		\sin\phi \\
		R_c \sin\phi \sin\varphi \\
		-R_c \sin\phi \cos\varphi \\
		0
	\end{pmatrix}
\end{equation}

\subsection{Computing 3D lines from a non-central projection}
\label{subsec:nocentralbbone}

Since it is a particular property we want to exploit, here we introduce how to compute a 3D line from a non-central projection system. Defining in Plücker coordinates a 3D line as $\mathbf{L} = (\mathbf{l}^T, \mathbf{\bar{l}}^T)^T \in \mathbb{P}^5$ (where $\mathbf{l} \in \mathbb{R}^3$ and $\mathbf{\bar{l}} \in \mathbb{R}^3$) and a projectng ray $\boldsymbol{\Xi} = (\boldsymbol{\xi}^T, \boldsymbol{\bar{\xi}}^T)^T \in \mathbb{P}^5$, their intersection is defined by the side operator \cite{pottmann2009computational} as:

\begin{equation} \label{eq:side}
	side(\mathbf{L},\boldsymbol{\Xi}) = \mathbf{l}^T \boldsymbol{\bar{\xi}} +
		\mathbf{\bar{l}}^T \boldsymbol{\xi} = 0
\end{equation}

Given that a 3D line has four degrees of freedom, we need, at least, 4 independent equations to solve the problem for $\mathbf{L}$. In general, four projecting rays from a 3D line generate four independent constraints from where we can compute the 3D line \cite{teller1999determining}. However, we can find some degenerate cases where four projecting rays do not generate independent constraints, e.g. the rays are coplanar with the revolution axis or with the plane containing the circle of optical centers.

\section{Layout estimation proposal}
\label{sec:proposal}

\begin{figure}
	\centering
	\includegraphics[width = 0.95\textwidth]{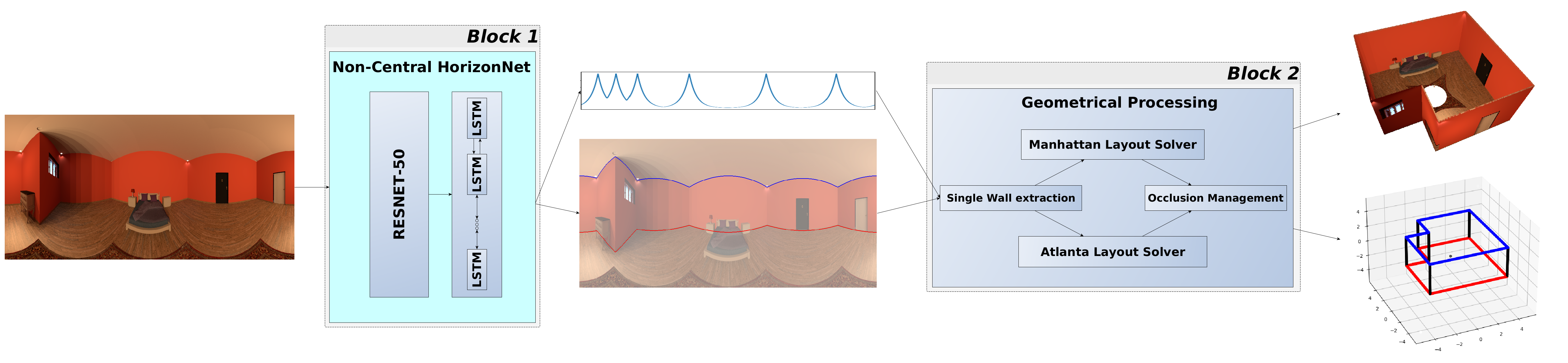}
	\caption{Pipeline of our proposal. In a first stage, the neural network extracts the boundaries of the structural lines of the room as well as a probability of corner positions from the non-central panorama. On a second stage, our proposed geometrical processing exploits the properties of non-central projection systems to recover the 3D of the layout from the information provided by the network.}
	\label{fig:pipeline}
\end{figure}

Our proposal for layout estimation is a new pipeline composed by two main blocks (see Fig. \ref{fig:pipeline}). In a first block, we use a neural network to obtain the boundaries of the structural lines of an indoor environment from an image. On the second block, we geometrically process the information provided by the network, exploiting the properties of non-central projection systems and recovering the scaled layout.

With respect to the first block, \cite{bermudez2016line} and \cite{zhang2014panocontext} propose geometrical methods based on hypothesis generation-verification to extract lines and layouts respectively while \cite{sun2019horizonnet} and \cite{pintore2020atlantanet} rely on the use of neural networks for layout recovery. 
In our proposal we combine both solutions in order to obtain the scaled layout from a single non-central panorama.
The use of a neural network allows to obtain the structural lines of an environment faster than with classical approaches of hypothesis generation-verification. In the next section we define in more detail the network architecture proposed and its advantages and disadvantages.
On the other hand, we aim to exploit the geometrical properties of non-central projection systems. In section \ref{sec:geometry} we present in detail the different geometrical solutions proposed to solve the layout reconstruction problem and the geometrical pipeline proposed to reconstruct the scaled layouts from single panoramas.

\subsection{Non-central HorizonNet}
\label{subsec:NCNet}

\begin{figure}
	\centering
	\includegraphics[width = 0.95\textwidth]{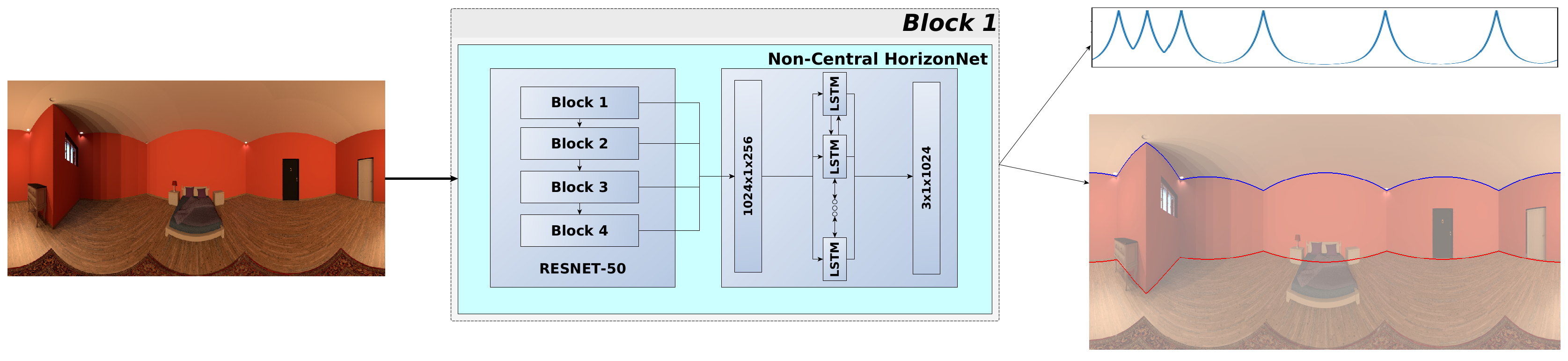}
	\caption{The non-central circular panorama is processed by Non-central HorizonNet, which is an adaptation of the work \cite{sun2019horizonnet}. First, it goes through a ResNet50, where high and low-level features are extracted. After a set of convolutions, the result is concatenated and fed to an array of bidirectional LSTMs. The network provides the boundaries of the structural lines of ceiling and floor, as well as the corners of the room as three separate 1D arrays.}
	\label{fig:pipeline_net}
\end{figure}

We propose to adapt an existing neural network in order to obtain the structural lines of indoor environments from non-central circular panoramas (see Fig. \ref{fig:pipeline_net}). This architecture is divided in two parts: a convolutional part, formed by the first layers of ResNet-50 \cite{he2016deep} and a set of convolutions; and a recurrent part, formed by a set of bi-directional LSTMs \cite{schuster1997bidirectional}. From this architecture, we obtain three 1D arrays with the boundaries information from the image. One of the arrays contains the probability of finding a wall-wall intersection in each column of the image. The other two 1D arrays provide the pixel of the intersection between the ceiling or the floor with the walls. From these three 1D arrays we obtain the boundaries of each of the structural lines that form the layout of the room.

The advantage of this architecture when dealing with non-central panoramas resides in how the network extracts the ceiling-wall and floor-wall intersections: column by column. Due to the bi-directional LSTMs, each column of the image is treated separately in order to recover the structural lines of the room. In our case, where non-central panoramas are used, this property is very interesting. As presented in section \ref{sec:noncentralmodel}, for each optical center, there are regions of the image that share the optical center. In particular, each column of the image is locally central, allowing the network to work in a central projection system for each separate column. Thus, this architecture proposed for central projection systems fits perfectly and is very suitable to extract the structural lines of a room from a non-central circular panorama.

HorizonNet imposes some restrictions that have to be considered to adapt it for non-central panoramas. The main restriction is that the image has to be oriented with the vertical direction. It means, that the wall-wall intersections form a straight vertical line in the image.  
Assuming this restriction, non-central panoramas must be acquired with the revolution axis of the system aligned with the gravity direction. This configuration introduces some disadvantages, since the depth and direction of lines parallel to the axis, the wall-wall intersection in this case, cannot be directly estimated (are one of the degenerated cases mentioned in section \ref{subsec:nocentralbbone}). However, since we know the gravity direction and that the structural lines will be perpendicular to it, we can turn the disadvantage into advantage, exploiting this constrait in the geometrical processisng in order to estimate the 3D lines of the layout. 

The original network architecture is trained in PanoContext \cite{zhang2014panocontext} and Stanford 2D-3D \cite{armeni2017joint}. These data-sets are formed by equirectangular panoramas obtained from indoor environments. In our proposal, we start with the network trained on these data-sets, since the distortion of equirectangular panoramas and non-central panoramas are similar. 
Afterwards we add a fine tunning to learn the particular distortion of the non-central panorama. For that purpose, we train the network, starting with the weights presented in \cite{sun2019horizonnet}, with a data-set formed of non-central panoramas and 3D information of the environment. However, since non-central projection systems are little used, there is no public data-set available. To solve this problem, we have generated and used a data-set of non-central circular panoramas from synthetic environments to fill this gap in the resources (more details in section \ref{subsec:dataset}). Once the network has been fine-tuned, it has learned the subtle different distortion of the non-central panoramas, providing more accurate information of the boundaries of the structural lines of the different environments.

\section{Geometrical processing}
\label{sec:geometry}

The next step in our proposal is to take advantage of the geometrical properties of the non-central panorama in order to recover the 3D layout and the scale of the environment. To do so, we propose a geometrical pipeline, which includes different linear solvers, that takes as input the information provided by the network and outputs the 3D corners of the room. 
In this section, we present the geometrical problem, defined as a plane extraction problem, and we provide two solutions to jointly obtain the whole layout of a room in Manhattan or Atlanta world assumptions. After these solutions, we present our detailed geometrical pipeline, that includes several steps to make more robust our implementation as well as handling occluded walls in the environment.

\subsection{Vertical wall extraction}
\label{subsec:problem}

\begin{figure}
	\centering
	\includegraphics[width = 0.45\textwidth]{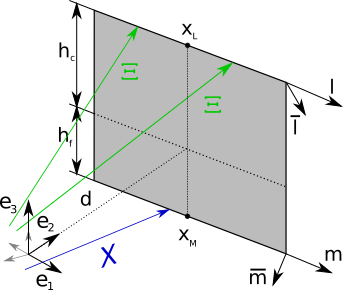}
	\caption{Rays and wall parameter definition. 
	The wall reference system is defined as $\{\mathbf{e}_1,\mathbf{e}_2,\mathbf{e}_3\}$; 
	$\boldsymbol{\Xi}$ and $\boldsymbol{X}$ are the projecting rays; 
	$\mathbf{(l,\bar{l})}$ and $\mathbf{(m,\bar{m})}$ are the ceiling and floor lines that define the wall;
	$\mathbf{x_L,x_M}$ are the closest points of the lines to the reference system; 
	$h_c$, $h_f$ and $d$ are the distance from the referende system to the ceiling, floor and wall planes respectively.}
	\label{fig:wallDef}
\end{figure}

Man-made environments are usually built by vertical walls that intersect the ceiling and the floor in two horizontal parallel straight lines. Thus, we define a wall as a set of two parallel horizontal lines $(\mathbf{L}, \mathbf{M})$ contained in a vertical plane (see Fig. \ref{fig:wallDef}). We define the ceiling line $\mathbf{L} = \left(\mathbf{l}^T,\mathbf{\bar{l}}^T \right)^T$ and the floor line $\mathbf{M} = \left(\mathbf{m}^T,\mathbf{\bar{m}}^T \right)^T$ in Pl{\"u}cker coordinates. 
We also define an orthonormal reference system placed in the origin and oriented with the vertical wall as $\{ \mathbf{e}_1,\mathbf{e}_2, \mathbf{e}_3 \} $. From the wall definition shown in Fig. \ref{fig:wallDef}, the lines direction coincide with the first component of the reference system $\mathbf{l} = \mathbf{m} = \mathbf{e}_1$. To define the momentum vector of the lines with respect to the reference system, we compute the cross product between the closest point of the line to the origin and the line direction as: $\mathbf{\bar{l}} = \mathbf{x_L} \times \mathbf{l}$ and $\mathbf{\bar{m}} = \mathbf{x_M} \times \mathbf{m}$, where $\mathbf{x_L}$ and $\mathbf{x_M}$ are the closest points of the lines to the origin. These points are defined as $\mathbf{x_L} = d \mathbf{e}_2 + h_c \mathbf{e}_3$ and 
$\mathbf{x_M} = d \mathbf{e}_2 + h_f \mathbf{e}_3$, where $h_c$, $h_f$ and $d$ refer to the distance from the reference system to the ceiling, floor and wall planes respectively.

\begin{equation}\label{eq:side_l_xi}
	side(\boldsymbol{\Xi},\mathbf{L}) = 
	\boldsymbol{\xi}^T \mathbf{\bar{l}} + \boldsymbol{\bar{\xi}}^T \mathbf{l} = 
	\boldsymbol{\xi}^T \left( h_c \mathbf{e}_2 - d \mathbf{e}_3 \right) + \boldsymbol{\bar{\xi}}^T \mathbf{e}_1 = 0
\end{equation}
\begin{equation}\label{eq:side_m_chi}
	side(\boldsymbol{X},\mathbf{M}) = 
	\boldsymbol{\chi}^T \mathbf{\bar{m}} + \boldsymbol{\bar{\chi}}^T \mathbf{m} = 
	\boldsymbol{\chi}^T \left( h_f \mathbf{e}_2 - d \mathbf{e}_3 \right) + \boldsymbol{\bar{\chi}}^T \mathbf{e}_1 = 0
\end{equation}

To compute the lines that define a wall from an image, we need the projecting rays of these lines. 
In our proposal, the neural network provides the pixel information of the boundaries of the projection of structural lines in the environment. From this pixel information, we can compute the projecting rays to the ceiling 
$\boldsymbol{\Xi} = \left( \boldsymbol{\xi}^T,\boldsymbol{\bar{\xi}}^T \right)^T$ 
and floor 
$\boldsymbol{X} = \left( \boldsymbol{\chi}^T,\boldsymbol{\bar{\chi}}^T \right)^T$ 
lines from the backprojection model seen in Section \ref{sec:noncentralmodel}. Known the projecting rays, we aim to obtain the 3D lines that define each wall. The relation among the projecting rays and the wall lines is given by their intersection, defined with the side operator in Plücker coordinates (see Section \ref{subsec:nocentralbbone}). From our definition of the wall, equations \eqref{eq:side_l_xi} and \eqref{eq:side_m_chi} define the intersection of the projecting rays and the walls of the layout. Trying to solve directly this ecuations may be difficult since, in general, is a non-linear problem. Instead, we propose a DLT-like \cite{abdel2015direct} approach where we compute the solution as a linear problem.

In a first approach, we aim to extract each wall of the layout independently. Let the main direction of a wall be horizontal and described by the vector $\mathbf{u} = (u_x,u_y)^T$ such that $\mathbf{l} = \mathbf{m} = (u_x,u_y,0)^T$. Then, the orthonormal reference system oriented with the wall can be re-defined as: 
$\{ \mathbf{e}_1,\mathbf{e}_2, \mathbf{e}_3 \} = \{(u_x,u_y,0)^T, (-u_y,u_x,0)^T, (0,0,1)^T\}. $
From this parameterization, equations \eqref{eq:side_l_xi} and \eqref{eq:side_m_chi} can be writen as:

\begin{equation}\label{eq:iter1}
	\bar{\xi}_1 u_x  + \bar{\xi}_2 u_y + \left(\xi_2 u_x - \xi_1 u_y \right)h_c - d \xi_3 = 0
\end{equation}

\begin{equation}\label{eq:iter2}
	 \bar{\chi}_1 u_x + \bar{\chi}_2 u_y + \left(\chi_2 u_x - \chi_1 u_y \right)h_f - d \chi_3 = 0
\end{equation}

These equations are non-linear since $\mathbf{u}$, $h_c$ and $h_f$ are coupled. 
At this point, we define the new variables $\mathbf{v} = h_c \mathbf{u}$ and $\mathbf{w} = h_f \mathbf{u}$. From this new variable definition, equations \eqref{eq:iter1} and \eqref{eq:iter2} become linear, obtaining the following equations:

\begin{equation}\label{eq:manx}
	\bar{\xi}_{1}  u_x + \bar{\xi}_{2}  u_y - \xi_{1}  v_y + \xi_{2}  v_x  - d \xi_{3}  = 0
\end{equation}
\begin{equation}\label{eq:many}
	\bar{\chi}_{1}  u_x - \bar{\chi}_{2}  u_y - \chi_{1}  w_y + \chi_{2}  w_x  - d \chi_{3}  = 0
\end{equation}

Now, we can build a linear system $\mathsf{A} \mathscr{W} = 0$, where the matrix $\mathsf{A}$ is full-filed with relations \eqref{eq:manx} and \eqref{eq:many} and $\mathscr{W} = (\mathbf{u}^T,\mathbf{v}^T,\mathbf{w}^T,d)^T$ is the unknown wall homogeneous vector. Notice that $\mathbf{u}$, $\mathbf{v}$ and $\mathbf{w}$ are independent variables which can be non-parallel. Since we have defined these vectors as proportional, to impose the parallelism we compute the null space of the system with a Singular Value Decomposition (SVD), obtaining a parametric solution which is a linear combination of singular vectors parameterized with $\lambda_i$. 
Notice that, two horizontal lines contained in a vertical plane have 4 dregrees of freedom. At this point we have two options to solve the problem. In one hand, a minimal solution would require 2 projecting rays for each line of the wall, describing the null space with three singular vectors and two parameters $\lambda_1$ and $\lambda_2$, such as $\mathscr{W} = {\mathscr{W}}_0 + \lambda_1 \mathscr{W}_1 + \lambda_2 \mathscr{W}_2$. By solving a system of two quadratic equations with action matrices or as a polynomial eigenvalue vector \cite{kukelova2011polynomial}, we obtain a set of 4 different solutions which should be discriminated. 
On the other hand, since the network provides enought robust information of the structural lines, we propose to solve the over-determined case, taking at least 3 rays for each line of the wall. 

\begin{equation}\label{eq:lambda_uv}
	\lambda(\mathbf{v_1} - h_c \mathbf{u_1}) = h_c \mathbf{u_0} - \mathbf{v_0}
\end{equation}
\begin{equation}\label{eq:lambda_uw}
	\lambda(\mathbf{w_1} - h_f \mathbf{u_1}) = h_f \mathbf{u_0} - \mathbf{w_0}
\end{equation}

In this over-determined case, the null space is described by a linear combination involving only one parameter $\lambda$, such as $\mathscr{W} = \mathscr{W}_0 + \lambda \mathscr{W}_1$. Imposing the parallelism restriction for $\mathbf{u}$, $\mathbf{v}$ and $\mathbf{w}$ as shown in equations \eqref{eq:lambda_uv} and \eqref{eq:lambda_uw}, we can derive two uncoupled quadratic equations for $\lambda$ as:

\begin{equation}\label{eq:quad_v}
\begin{split} 
	(u_{y0} u_{x1} - u_{x0} u_{y1}) \lambda^2 &\\
	+ (u_{x0} v_{y1} + v_{x0} u_{y1} - u_{y0} v_{x1} - v_{y0} u_{x1} ) \lambda &\\
	+ (v_{y0} v_{x1} - v_{x0} v_{y1})& = 0
\end{split}
\end{equation}

\begin{equation} \label{eq:quad_w}
\begin{split}
	(u_{y0} u_{x1} - u_{x0} u_{y1}) \lambda^2 & \\
	+ (u_{x0} v_{y1} + w_{x0} u_{y1} - u_{y0} w_{x1} - w_{y0} u_{x1} ) \lambda & \\
	+ (w_{y0} v_{x1} - w_{x0} w_{y1}) & = 0
\end{split}
\end{equation}

Computing the solution for $\lambda$ in each equation, we get four solutions. However, the solutions from equation \eqref{eq:quad_v} and equation \eqref{eq:quad_w} are paired, which means that efectively we have only two different solutions for $\lambda$. The global orientation prior allows to easily discriminate which of the solutions is the correct one. Computing $\mathscr{W}$ for each $\lambda$ and extracting the ceiling $h_c$ and floor $h_f$ plane distances, we observe that only one of the solutions sets $h_c > h_f$. Taking the correct value of $\lambda$, we have defined the wall direction $\mathbf{u}$, the ceiling $h_c$, floor $h_f$ and wall $d$ planes distance to the acquisition system as well as the Plücker coordinates of the ceiling and floor lines that define the wall.

\subsection{Manhattan layout solver}
\label{subsec:manhattanDLT}

Notice that in a Manhattan world assumption, there is a set of walls sharing the wall direction $\mathbf{u} = (u_x,u_y)^T$ and the complementary set of walls share the orthogonal direction $\mathbf{u}_\perp = (-u_y,u_x)^T$. Since we assume that the rooms have single ceiling and floor planes, all the walls share the ceiling $h_c$ and floor $h_f$ heights. Defining the projecting rays of the walls with direction $\mathbf{u}_\perp$ as $(\boldsymbol{Z},\boldsymbol{\Psi})$, we can redefine the equations \eqref{eq:manx} and \eqref{eq:many} for this set of walls as:

\begin{equation}\label{eq:manPx}
	\bar{\zeta}_{2}  u_x - \bar{\zeta}_{1}  u_y - \zeta_{2}  v_y - \zeta_{1}  v_x  - d_i \zeta_{3}  = 0
\end{equation}
\begin{equation}\label{eq:manPy}
	\bar{\psi}_{2}  u_x - \bar{\psi}_{1}  u_y - \psi_{2}  w_y - \psi_{1}  w_x  - d_i \psi_{3}  = 0
\end{equation}

Then, we can extend the DLT-like approach to fit all the set of walls computing the null space of $\mathsf{A} \mathscr{L}_M = 0$, where $\mathscr{L}_M = (\mathbf{u}^T,\mathbf{v}^T,\mathbf{w}^T,d_1,\cdots,d_N)^T$, where $N$ is the number of walls, and the matrix $\mathsf{A}$ is full-filed with relations \eqref{eq:manx} and \eqref{eq:many} for a set of walls and \eqref{eq:manPx} and \eqref{eq:manPy} for the other.
This approach allows to jointly obtain the main directions in the Manhattan world assumption, the height of the room and the walls locations.

\subsection{Atlanta layout solver}
\label{subsec:globalDLT}

In the case of Atlanta world assumption, each wall of the room could have a different horizontal direction, therefore we must find a new approach to obtain the whole layout. Notice that from the previously proposed solver, we can extract each wall independently. However, with this approach we do not impose that the walls share the ceiling and floor heights. Nevertheless, if the direction of each wall is known (e.g. extracting each wall independently), we can derive a new linear solution for the whole layout.

\begin{equation}\label{eq:genx}
	\bar{\xi_1}' + h_c \xi_2' - d \xi_3' = 0
\end{equation}
\begin{equation}\label{eq:geny}
	\bar{\chi_1}' + h_f \chi_2' - d \chi_3' = 0
\end{equation}

\begin{figure}
	\centering
	\includegraphics[width = 0.95\textwidth]{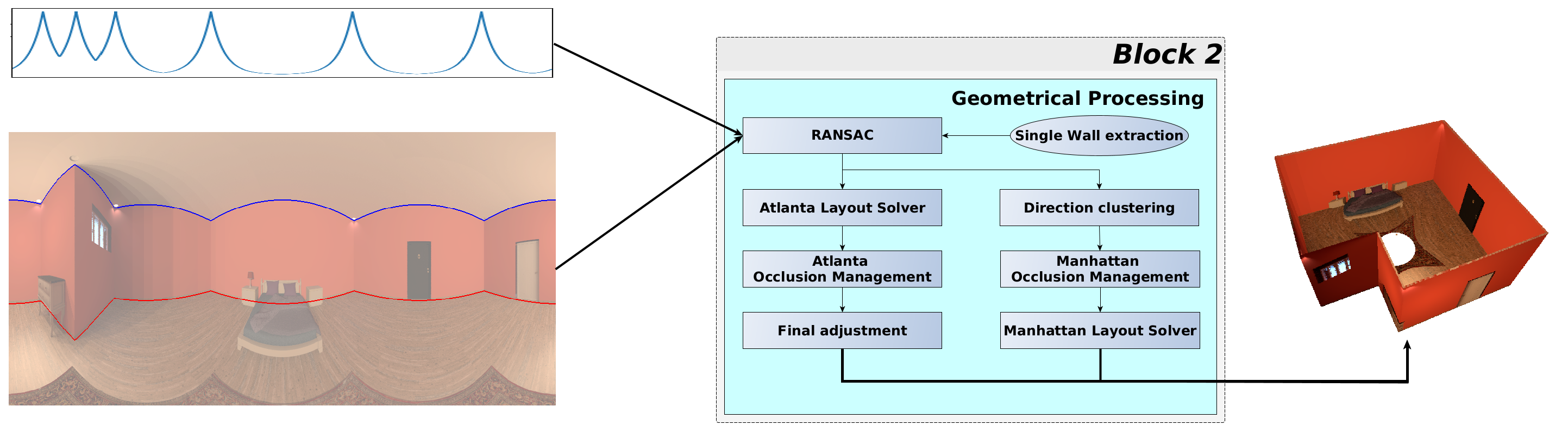}
	\caption{Geometrical pipeline. The input is the pixel information provided by the network. First, a Ransac makes a first wall direction estimation. Then the pipeline branches. For Manhattan world assumption, we cluster the walls direction, then handle the possible occlusions and compute the direction labels of the walls. Finally we compute the Manhattan layout with the proposed solver. For Atlanta world assumption, once defined the walls directions, we implement the proposed layout solver. Then we search for occlusions. As last step, a final adjustment of the corners is made to obtain the final 3D layout.}
	\label{fig:pipeline_geom}
\end{figure}

Making a first independent wall extraction, we obtain the direction for each wall. Changing the refence system of the projecting rays from the acquisition system to each wall local reference system, equations \eqref{eq:side_l_xi} and \eqref{eq:side_m_chi} become the linear expresions \eqref{eq:genx} and \eqref{eq:geny} respectively, where $\boldsymbol{\Xi}'$ and $\boldsymbol{X}'$ are the projecting rays in each wall reference system. Then we can solve the null space of the linear system $\mathsf{A} \mathscr{L}_A = 0$, where $\mathsf{A}$ is a matrix full-filed with equations \eqref{eq:genx} and \eqref{eq:geny} and $\mathscr{L}_A = (1,h_c,h_f,d_1,\cdots,d_N)^T$, where $N$ is the number of walls in the environment.
In this approach, once known the walls directions, we can simultaneously compute the room height and the walls location. 

\subsection{Detailed Geometric Pipeline}
\label{subsec:pipeline}

In order to improve the robustness of our method, we propose a new full geometric pipeline that includes the two new solvers presented before (see Fig. \ref{fig:pipeline_geom}). This pipeline takes as input the information provided by the network and gives as output the 3D lines and corners that form the layout. 
This geometric pipeline is divided in two branches, one for Manhattan world assumption and other for Atlanta world. This is due to the different management of the occlusions in each world assumption.

The geometric pipeline starts with a RANSAC algorithm to filter possible spurious data. Here could rise a question: 
What is the advantage of using these structural deep learning based edges and corners over classic Canny edges if we still have to use a RANSAC approach? The main advantage is the huge reduction of the number of required hypotheses. Consider the number of hypotheses in a RANSAC approach $n_{hyp}$ which is typically estimated by 
$n_{hyp} = \frac{\log(1-P)}{\log(1-(1-\epsilon)^k)} $,
where $P$ is the probability of not failing in the random search, $\epsilon$ is the rate of outliers and $k$ the number of elements defining the hypothesis.
Assuming that we want to extract a wall, with a probability of $P=99.99\%$ of not failing, we need two lines, defined by 3 points each. We assume a rate of outliers of $\epsilon=20\%$ in the input data. 
With our method, we have well defined which data belong to the ceiling line and which to the floor line and the data of each wall separately. So, we assume that our $\epsilon=20\%$ and that we only need $k=3$ samples to define the two lines, since data is defined by column and we can take one value for each line in each column. This computation leaves that our implementation needs $n_{hyp} = 12,84$ hypotheses to define the best wall that fits the data. 
With state-of-the-art approaches as \cite{bermudez2016line,bermudez2017exploiting}, where an edge detector is used, as Canny, the data is not as well defined and structured. Thus, the probability of a sample been an inlier is reduced by half, since it can be part of the ceiling or the floor. This also means that we need the double of samples, since they are not paired. This assumptions lead to a number of outliers of $\epsilon=60\%$ and a number of samples $k=6$, obtaining $n_{hyp} = 2244$ hypotheses per wall. This difference in the number of hypotheses is reduced by the neural network.
Thus, with the solution for a single wall presented in section \ref{subsec:problem} as hypothesis in the RANSAC, we get the 3D lines that better fit the information provided by the network. After this step, the pipeline branches depending on the world assumption.

{\bf Assuming a Manhattan world}. We cluster the extracted walls into two classes corresponding to the main perpendicular directions. In this clustering, we label each wall with the index of one of the clusters corresponding to different perpendicular Manhattan directions.
In a second step, we manage the occlusions in the Manhattan environment. Since a Manhattan world only have two main directions, these must be alternating in consecutive walls. If two consecutive walls have the same direction means that an occluded wall is between them. In this case, we add a perpendicular wall between the occluded and occluder walls to keep the alternation in the walls direction. 
Finally, once we know the number of walls and their Manhattan direction label (defined as $\mathbf{u}$ and $\mathbf{u_\perp}$ in Section \ref{subsec:manhattanDLT}) we apply the Manhattan layout solver. This solver will provide the walls direction that better fit the whole environment as well as the height of the room and the walls location. Once obtained the lines that define the walls, we can obtain the 3D corners of the room computing the intersection of these lines, which is easy since ceiling lines are co-planar, as well as floor lines.

\begin{figure*}[t]
	\centering
	\includegraphics[width = 0.95\textwidth]{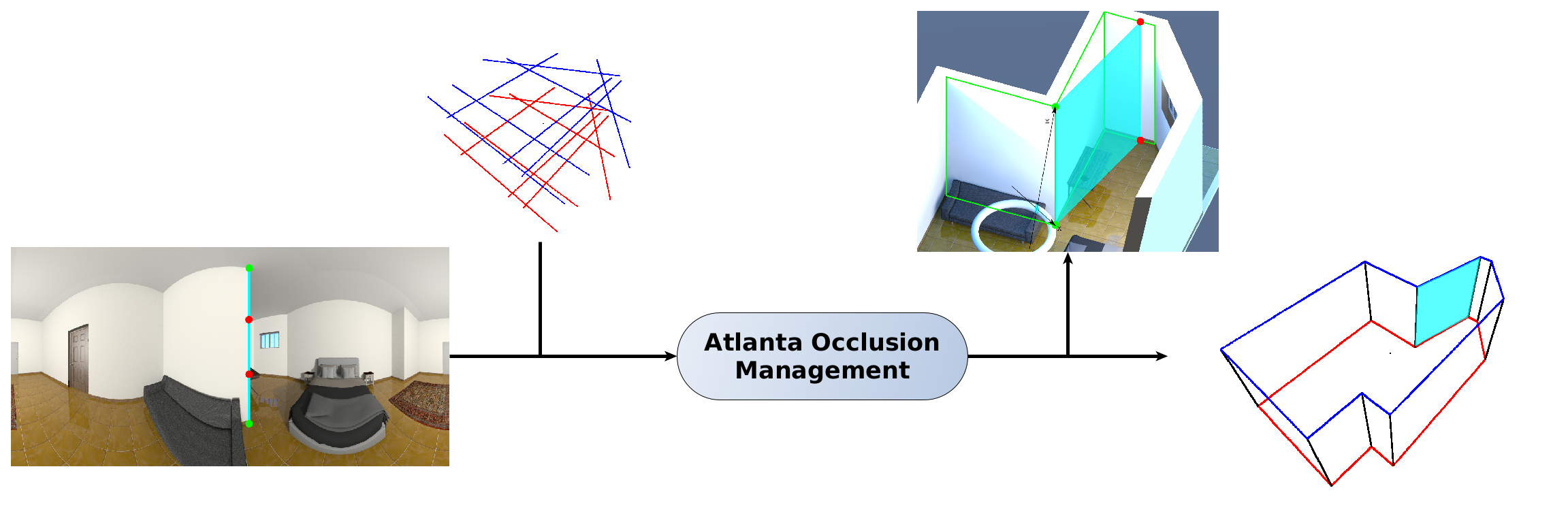}
	\caption{Atlanta occlusion management. We generate a new wall (in blue) which is co-planar with the projecting rays $\boldsymbol{\Xi}$ and $\boldsymbol{X}$ correspondng to the visible corner (dots in green). This new wall defines two new corners (dots in red) in a partialy occluded wall. The green wireframe is the ground truth layout.}
	\label{fig:Atocc}
\end{figure*}

{\bf Assuming an Atlanta world}. We do not know the number of dominant directions in an Atlanta environment. Besides, we cannot find an occlusion looking at the walls direction, since two consecutive walls may have a similar orientation. Thus, Atlanta enviroments must be tackled in a different way than the Manhattan ones.

Since Atlanta environments do not have a defined number of main directions, we cannot cluster the walls extracted in the RANSAC by their direction. By contrast, we consider this first wall direction estimation good enough and use it as initial value in our pipeline.
Then, assuming that the walls direction is known, we can apply the solution for Atlanta environments presented in section \ref{subsec:globalDLT}, where we jointly obtain the room height and the walls locations.

Once defined the lines that form the walls of the environment, we compute the 3D corners. Notice that if we compute the corners as lines intersection, we may make impossible layouts if there is an occlusion in the environment. To avoid this problem, we compute each corner as the point of the computed 3D line that minimises the Euclidean metric distance in $\mathbb{E}^3$ \cite{bermudez2017exploiting} between the 3D line and the projecting ray corresponding to the corner provided by the network. 
Each projecting ray crosses with two 3D lines corresponding to two consecutive walls where two different cases may appear. When the computed 3D corners coincide in a single point, no occluded wall is detected and this point is a corner of the environment. By contrast, if the computed corners are different 3D points (i.e. the distance between them is higher than a threshold), we have found an occluded wall between the two 3D lines and we insert a wall in the layout model (see Fig. \ref{fig:Atocc}). Since in an Atlanta world assumption there is no restriction about the walls direction, we assume that the occluded wall and the projecting ray of the corner lie in the same plane. 

Finally, we make a final adjustment where we fine tune the 3D lines direction and position. We minimise the least-square reprojecting error of the computed 3D lines with the pixel coordinates of the boundaries \cite{bermudez2017exploiting} provided by the network. This final adjustment refines the position of the 3D corners. The movement of the corners out of the plane defined by the projecting rays of the corners is penalized. This extra soft-restriction allows managing the infinite posible solutions caused by occlusions during the optimization step. 

\section{Experiments}
\label{sec:experimets}

The pipeline of our proposal is divided into two main blocks: a neural network that extracts the boundaries of the structural lines of an indoor environent from a single panorama and a geometrical processing that takes as input the output of the network and recovers the 3D scaled layout. In order to evaluate our proposal, we have performed a set of experiments. 
We independently evaluate the performance of both main blocks: the proposed neural network and geometric pipeline. 
Additionally, we make a comparison with state of the art methods for line estimation in non-central panoramas and for layout recovery from single panoramas.
We also present qualitative results of our method. In Figure \ref{fig:results} we show examples of the layout reconstruction from virtual environments while in Figure \ref{fig:realExp} we present examples with real images.
Before describing the experiments, we present the image dataset used to train the neural network and perform the experiments.

\subsection{Non-central panorama data-set}
\label{subsec:dataset}

Currently, we can find a great amount of image data-sets, from perspective images \cite{deng2009imagenet} to omnidirectional panoramas \cite{armeni2017joint,chang2017matterport3d}. However, non-central projection systems have never been used with deep learning before, thus there is not a data-set of this kind of images. This is a big problem in our case, since we need a large amount of images to train a deep learning architecture. So, we present a new dataset obtained with our synthetic generator of realistic non-central panoramas. It includes semantic, depth and 3D information of the environment.

We generate random layouts, from 4 to 14 walls, in a Manhattan world assumption. Then, with a probability set by the user, corners of these layouts are clipped and substituted by oblique walls to generate Atlanta world layouts. Once obtained the structure of the room, we compute the free space to place objects in it. We have two kinds of objects: those that are placed next to a wall in a fixed orientation (beds, desks, wardrobes, TVs) and those that are placed in the middle of the room at any orientation (chairs, sofas, carpets). These objects are taken from different pools where one is chosen randomly and placed in the room if has a free space in it. Additionally, the ceiling, floor and walls materials and colour are taken from different pools and chosen randomly for each new environment. After the virtual environment is generated, we set the illumination conditions. We have different pre-defined ambient illuminations and we also randomly place spot lights (ceiling lamps) to give the environment a more realistic view.

Once defined the virtual enviroment, we use POV-Ray\footnote{The Persistence of Vision Raytracer \url{http://www.povray.org/}, website visited in 2020} to render RGB and semantic images and MegaPOV\footnote{MegaPOV \url{http://megapov.inetart.net/}, website visited in 2020} for depth maps. These images are generated by a ray tracing method, which can follow ad-hoc programmable camera projection models. In our case, we use the projection model presented in Section \ref{sec:noncentralmodel} for non-central circular panoramas. The center of the acquisition system is placed in a random position for each room generated, obtaining a greater variability in the walls distortion along the data-set. This allows to generate a set of images, from different rooms and in different positions, not only in the center of the room as many existing data-sets.

For this work, we have generated a data-set of non-central panoramas to fine-tune the deep learning architecture presented in section \ref{subsec:NCNet}. The data-set\footnote{The data-set will be available under request in {github.com/jesusbermudezcameo/NonCentralIndoorDataset}} is formed by more than \textbf{2600} images, taken from different positions inside the environments,  from around \textbf{650} different rooms, from 4 to 14 walls, combining Manhattan and Atlanta environments. We propose a division of the dataset in three blocks: training set, formed by 1677 panoramas; validation set, formed by 399 panoramas; and test set, formed by 499 panoramas. Each set includes Atlanta and Manhattan environments of different number of walls, and we make sure that there are no equal layouts in different sets. 

\subsection{Ablation study: Non-central HorizonNet}
\label{subsec:network_exp}

In this work, we have adapted an existing neural network to work with non-central panoramas. We have fine-tunned the network with a new data-set of non-central panoramas from Manhattan and Atlanta indoor environments. We take the weights for the network that minimise the validation error to perform the different experiments. 

We first evaluate the performance of the network. In order to evaluate how well the network obtains the boundaries of the structural lines and the corners of indoor environments, we evaluate our fine-tunned network against the test-set of our dataset and compare it against the original weights of \cite{sun2019horizonnet}. 

To evaluate the performance of the boundary extractor part of the network, we will compute the pixel error of the output. Since the network provides two 1D arrays with the pixel coordinates of the boundaries of the structural lines, we compute the pixel distance (PE) for each image column. 
In the case of the corner prediction, we compare the corner output of the network, which is a probability distribution, with the distance function defined as label for the network. The distance function used is the same as in \cite{sun2019horizonnet}, which is defined as $y_{cor} = c^{d_x}$, where $y_{cor}$ is the function value, $c$ is a constant value set to $c=0.96$ and $d_x$ is the pixel distance to the closest corner.
The metrics defined for the corner prediction evaluation are: Precision, Recall, Accuracy and Intersection over Union (IoU). In Table \ref{tab:netcorExp} we present the results of our trained network tested in the proposed data-set.

\begin{table*}
\centering
\caption{Network evaluation and comparison against \cite{sun2019horizonnet} with non-central panoramas. For the corner extraction evaluation, metrics are defined up to 1, being greater numbers better. For the boundary extraction, smaller numbers are better. Both networks are tested in the test partition of the dataset proposed in section \ref{subsec:dataset}.}
\label{tab:netcorExp}
\begin{tabular}{l|cccc|cc|}
			\cline{2-7}
&	\multicolumn{4}{c|}{Corner prediction}			&	\multicolumn{2}{|c|}{Boundary Extraction}			\\ \cline{2-7}
						&	Precision	&	Recall	&	Accuracy	&	IoU		& PE (mean)		&	PE (std)\\ \hline
\multicolumn{1}{|l|}{HorizonNet \cite{sun2019horizonnet}}	
						&	0.797		&	0.445	&	0.912		&	0.407	&	10.766		&	5.271	\\ \hline
\multicolumn{1}{|l|}{Non-central}	
						&	\multirow{2}{*}{0.934}		&	\multirow{2}{*}{0.962}	&	\multirow{2}{*}{0.986}		&	\multirow{2}{*}{0.901}
						&	\multirow{2}{*}{1.167}		&	\multirow{2}{*}{0.490}	\\ 
\multicolumn{1}{|c|}{HorizonNet}& & & & & & \\ \hline
 \multicolumn{1}{c}{}& \multicolumn{4}{c}{\it higher is better} & \multicolumn{2}{c}{\it lower is better} 
\end{tabular}
\end{table*}

\subsection{Ablation study: geometric solvers}
\label{subsec:geometry_exp}

We evaluate the proposed solvers presented in Section \ref{sec:geometry}. To do so, we use the ground truth information of the test partition of the dataset to evaluate the sensitivity to noise of the geometric solvers for line extraction. We compare our results with the state of the art method for line extraction in non-central panoramas \cite{bermudez2016line}. 
Since we are focusing on the geometric approach, we omite the environments with occlusions.

We use as input information the projecting rays of the boundaries of the structural lines, with sub-pixel precision, of the indoor environment, taken from the ground truth information of the data-set, and the real wall directions for each wall in the layout. To evaluate the sensitivity to noise of the solvers, we add increasing Gaussian noise to the ground truth projecting rays at sub-pixel level. For the evaluation of our Manhattan layout solver, we use the walls direction to label the walls in the two main Manhattan directions. For the Atlanta layout solver, we need these wall directions to compute the projecting rays in the wall reference system. The method proposed in \cite{bermudez2016line} compute the lines which are parallel to a known plane, which in our case is the horizontal plane.

To make the evaluation and comparison, we use the same metrics defined in \cite{bermudez2016line}. Computed a 3D line $\mathbf{L} = (\mathbf{l},\mathbf{\bar{l}})^T$, we compute the direction error as: 
$\epsilon_{dir} = \arccos (\mathbf{l}^T \cdot \mathbf{l}_{gt})$, measured in degrees. 
We also compute the depth error of the line as: 
$\epsilon_{depth} = | \| \mathbf{\bar{l}}\| - \| \mathbf{\bar{l}}_{gt}\| |$, measured in meters.
Additionally, we use a common metric in layout estimation works, the corner error (CE). We compute the corners of the layout as line intersections and compute the L2 distance to the ground truth corners to define the metric.

\begin{figure}
	\centering	
	\subfloat[]{\includegraphics[width=0.32\textwidth, valign=c]{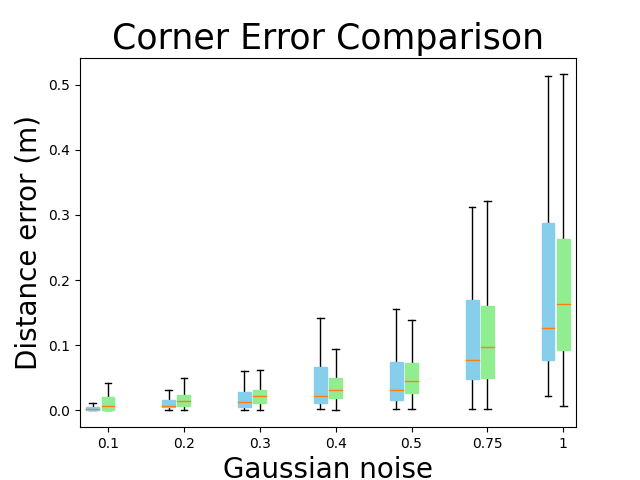}} \hfil
	\subfloat[]{\includegraphics[width=0.32\textwidth, valign=c]{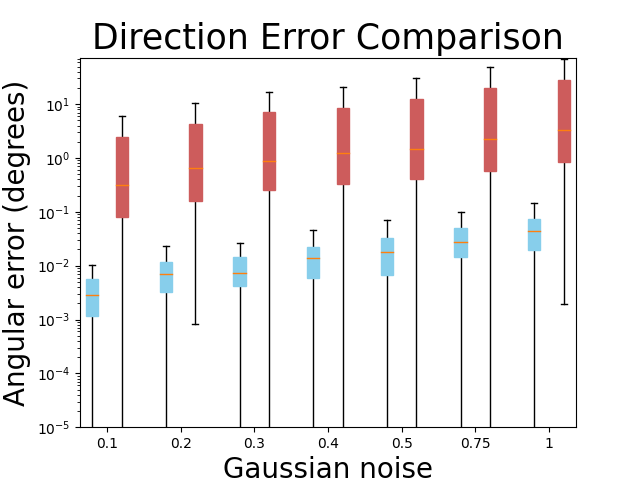}} \hfil
	\subfloat[]{\includegraphics[width=0.32\textwidth, valign=c]{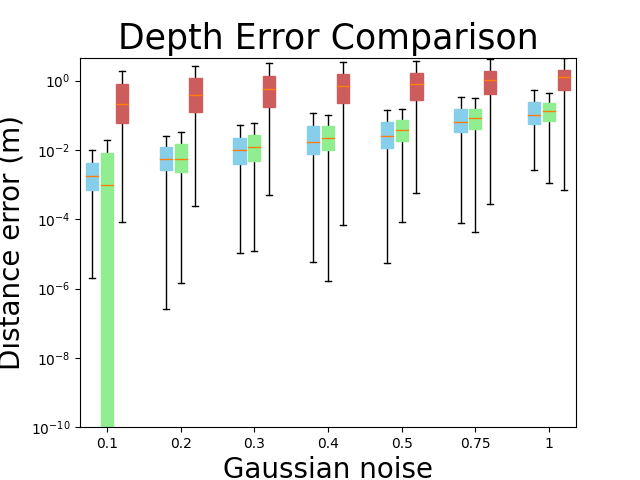}} 
	\caption{Experimental results to evaluate and compare our proposed Manhattan and Atlanta layout solvers against a state-of-the-art method for line extraction \cite{bermudez2016line}. 
	In blue are the Manhattan solver results.
	In green are the Atlanta solver results.
	In red are the state of the art method results.
	(a) shows the Corner error, in meters, of our proposed solvers. 
	(b) shows the Direction error of our solvers compared to \cite{bermudez2016line}, in degrees and logarithmic scale. Notice that the direction error of the Atlanta solver is zero since we provide the lines direction as prior. 
	(c) shows the Depth error of our proposed solvers compared to \cite{bermudez2016line}, in meters and logarithmic scale. }
	\label{fig:Geomcomparison}		
\end{figure}

The evaluation and comparison of our methods with the proposed in \cite{bermudez2016line} is shown in Figure \ref{fig:Geomcomparison}.
In blue is shown our Manhattan solver, in green is shown our Atlanta solver while in red is shown the method presented in \cite{bermudez2016line}. 

\subsection{Full pipeline validation}
\label{subsec:pipe_exp}

In this section, we analyse the performance of the geometric pipeline proposed in section \ref{subsec:pipeline} against the use of only the geometric solvers from sections \ref{subsec:manhattanDLT} and \ref{subsec:globalDLT}. The question is: if we have two geometric solvers that obtains the whole layout, why we need a more complex geometric pipeline? The answer comes in two parts. First, the geometrical solvers are not able to handle possible occlusion in the environment. Second, we have observed that the input information to the geometric block of our full pipeline (Fig \ref{fig:pipeline}) can be noisy and with spurious data. As seen in previous section \ref{subsec:geometry_exp}, the geometrical solvers work well with refined data, however they may lead to impossible layouts if the input is too noisy or if the fitting is corrupted by spurious information. 

We test how much the performance of our method improves with and without the proposed geometric pipeline. For that purpose, we compute the corners of the test-set layouts of our dataset in two different cases. In the first case, we use as input of our geometric block the labels of the network  ('Network labels' in table \ref{tab:pipeExp}), that means, the best information that the network would be able to provide. On a second case, we use as input of the geometric block the predictions provided by the network ('Network predictions' in table \ref{tab:pipeExp}), which are noisier than the labels. The metrics used to make the comparison are: Corner Error (CE), defined as the L2 distance between the computed corners and the ground truth; and the Intersection over Union (IoU) of the volumes of the computed layout and the groudn truth. In this experiment , 'Solvers' refer to the use of only the geometric solvers presented in section \ref{sec:geometry}. We use the wall extractor (see section \ref{subsec:problem}) to compute the wall labels in the Manhattan case and the walls' direction in the Atlanta case, and then the layout solver to obtain the corners of the room. 'Pipeline' refers to the use of the whole geometric pipeline proposed in section \ref{subsec:pipeline}. In order to make a more fair comparison, environments with occlusions have not been taken into account to compute the metrics (if so, the performance difference would be much greater). Table \ref{tab:pipeExp} shows the results of this experiment.

\begin{table*}
\centering
\caption{Comparison of layout extraction with only the solvers or the whole proposed geometric pipeline.}
\label{tab:pipeExp}
\begin{tabular}{lc|cc|cc|}

\cline{3-6}
 & & \multicolumn{2}{|c|}{Network labels} & \multicolumn{2}{|c|}{Network predictions} \\ \cline{2-6}
 & \multicolumn{1}{|c|}{World assumption} & CE (m) & IoU (\%)  & CE (m) & IoU (\%) \\ \hline
\multicolumn{1}{|l|}{\multirow{2}{*}{Solvers}}
								& Manhattan & 0.0297  & 98.0036  & 0.5569  & 81.7364 \\
\multicolumn{1}{|l|}{}			& Atlanta	& 0.8575  & 57.3289  & 1.2860  & 42.6701 \\ \hline
\multicolumn{1}{|l|}{\multirow{2}{*}{Pipeline}}
								& Manhattan & 0.0218  & 98.4753 & 0.2109  & 86.8104  \\
\multicolumn{1}{|l|}{}			& Atlanta	& 0.1391  & 92.5012 & 0.4811  & 76.0218  \\ \hline

\end{tabular}
\end{table*}

\subsection{State of the art comparison}
\label{subsec:comparison}

We have performed two different tests to compare our work with the state-of-the-art methods for layout reconstruction from a single panorama. In a first experiment, we have fine-tuned HorizonNet \cite{sun2019horizonnet} with equirectangular panoramas of the same virtual environments of our data-set. On a second experiment, we compare the results of different state-of-the-art works with ours. 

For our first experiment, we have generated a dual test-set, with equirectangular and non-central panoramas taken in the same position in the same Manhattan environments. We use this set-up to compare two different methods and two different acquisition systems in the same test-set. Then, we reconstruct the layout with HorizonNet, from central panoramas, and with our method, from non-central panoramas, for Manhattan environments. Notice that HorizonNet does not extract the scale of the layout. Instead, it assumes a camera height and computes the 3D corners with this extra measurement. Our method does extract the scale of the environment, so this measurement is computed and not given. To take this into account, we compare the metrics in two different cases. In a first case (No-scale), we normalize the layouts, predicted and ground truth, with the camera height. In the second case (Metric), we use the real camera height to compute the scale in the prediction from HorizonNet, leaving our method as has been presented in this work.

Table \ref{tab:fairComparison} shows the results of this comparison. The metrics used are the Corner Error (CE) and the Intersection over Union (IoU) defined before. We observe that our proposed method has better performance than HorizonNet, even thought both use the same network architecture.

On a second experiment, we compare our method with other state-of-the-art methods. This second comparison is not as fair as the previous one since each method has been trained and tested on different data-sets, so the results can depend on the dataset used and not only on the method. Besides, as stated before, our proposal only uses the image information in order to recover scaled 3D layout while the rest of the methods in the state of the art provide up-to-scale measures, relying on some extra measurement in the environment for the 3D reconstruction (e.g. the camera height).
The metrics used for the comparison are: 3D IoU, which refer to the 3D intersection over union of the predicted layout and the ground truth; 3D IoU(u2s), which refer to the up-to-scale intersection over union of the layout; CEN, which refer to the Corner Error Normalized computed as the L2 distance of the corners divided by the diagonal of the layout's bounding box; CE, which refers to the Corner Error computed as the L2 distance of the corners in meters. Table \ref{tab:comparison} shows the results of this experiment.

\begin{table}
\centering
\caption{Comparison of methods: fine tuned HorizonNet \cite{sun2019horizonnet} (HorizonNet FT) and our method. Evaluation made in selected Manhattan environments from the test-set of the presented dataset.}
\label{tab:fairComparison}
\begin{tabular}{lc|cc|}

\cline{2-4}
 & \multicolumn{1}{|c|}{Scale assumption} & CE  & IoU (\%)\\ \hline
\multicolumn{1}{|l|}{\multirow{2}{*}{HorizonNet FT}}
								& No-scale  & 0.3380  & 82.6742 \\
\multicolumn{1}{|l|}{}			& Metric	& 0.3504  & 82.6742 \\ \hline
\multicolumn{1}{|l|}{\multirow{2}{*}{\bf Ours}}
								& No-scale  & 0.2024  & 87.6208 \\
\multicolumn{1}{|l|}{}			& Metric	& 0.2271  & 87.6208 \\ \hline

\end{tabular}
\end{table}

\begin{table}
\centering
\caption{Comparison of different state-of-the-art methods for 3D layout recovery.}
\label{tab:comparison}
\begin{tabular}{l|cc|cc|}

\cline{2-5} 
 & \multicolumn{4}{|r|}{Manhattan World assumption} \\ \cline{2-5}
                   &   3D IoU (u2s) & 3D IoU    &	CEN    &	CE \\ \hline
\multicolumn{1}{|l|}{CFL \cite{fernandez2020corners}}          
                        &	78.87	    & -     &	0.75	&	-	     \\
\multicolumn{1}{|l|}{HorizonNet \cite{sun2019horizonnet}}   
                        & 	82.66	    & -     &{\bf 0.69}	&	-	     \\
\multicolumn{1}{|l|}{AtlantaNet \cite{pintore2020atlantanet}}   
                        &	83.94	    & -     &	0.71	&	-	     \\
\multicolumn{1}{|l|}{\bf Ours }     
                        &   {\bf 93.87 }& {\bf 86.16} &	0.78&  {\bf 0.223}     \\ 
\hline
 & \multicolumn{4}{|r|}{Atlanta World assumption}\\
\cline{1-5} 
\multicolumn{1}{|l|}{HorizonNet \cite{sun2019horizonnet}}   
                                    & 	73.53	    & -     &	-   	&	-	     \\
\multicolumn{1}{|l|}{AtlantaNet \cite{pintore2020atlantanet}}   
                                    &	80.01	    & -     &	-		&	-	     \\
\multicolumn{1}{|l|}{\bf Ours }     & {\bf 90.46}   & {\bf 76.02} &	{\bf 1.5}  &  {\bf 0.481}     \\ 
\hline
 \multicolumn{1}{c}{}& \multicolumn{2}{c}{\it higher is better} & \multicolumn{2}{c}{\it smaller is better} 
\end{tabular}
\end{table}

\subsection{Qualitative experiments: Real images}
\label{subsec:realimg}

\begin{figure}
\centering
	\subfloat{\includegraphics[width=0.3\textwidth ,valign=c]{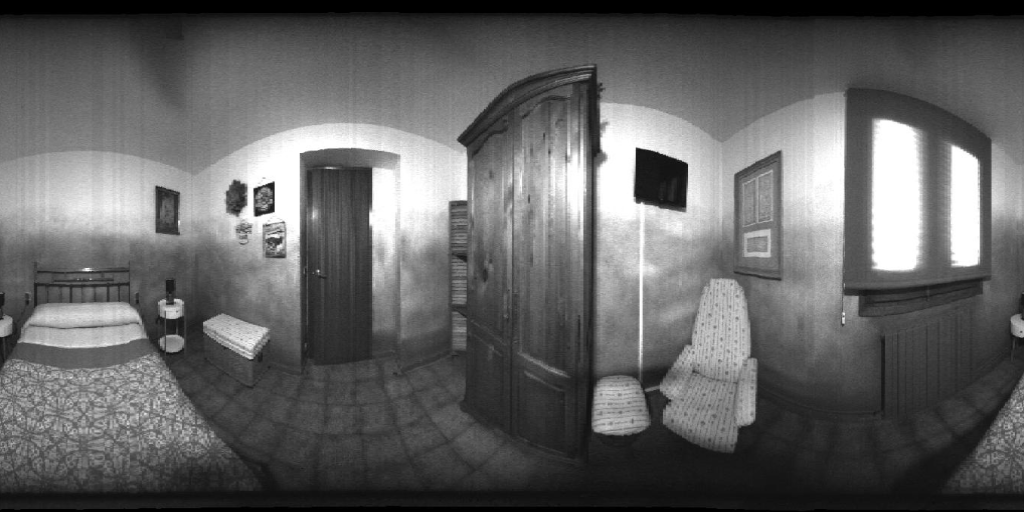}} \hfil
	\subfloat{\includegraphics[width=0.3\textwidth ,valign=c]{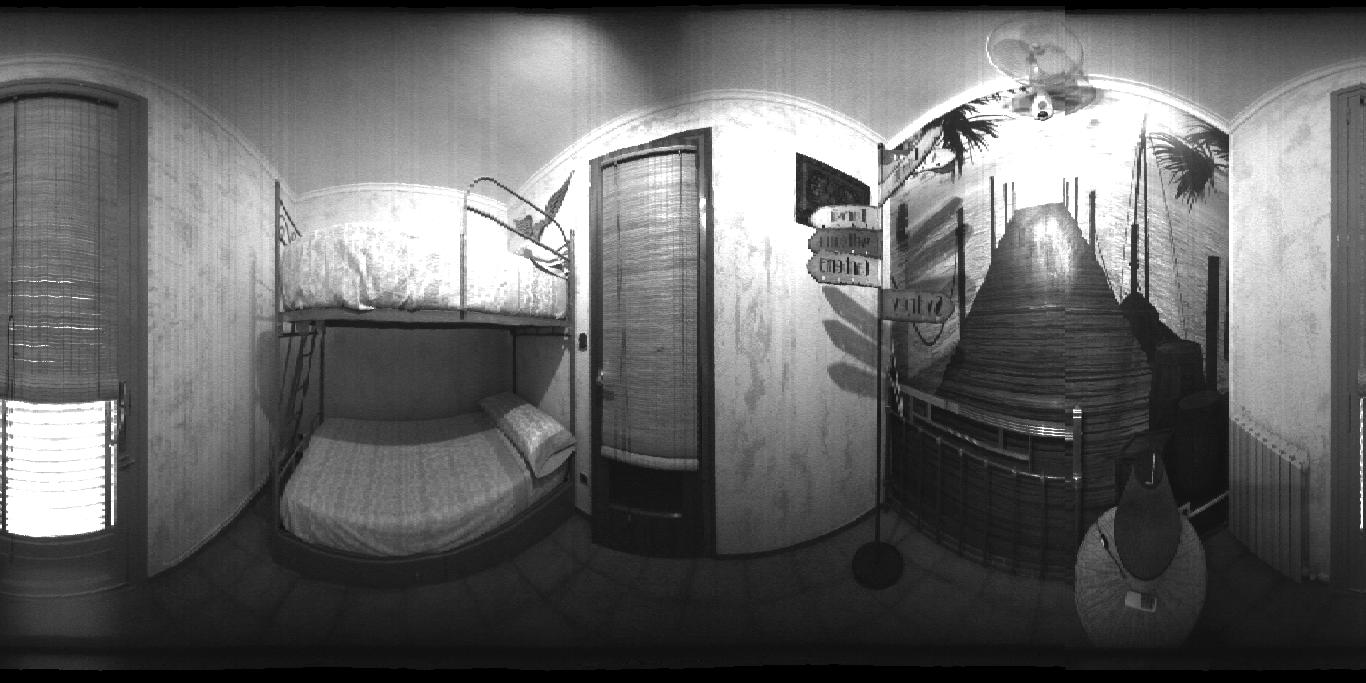}} \hfil
	\subfloat{\includegraphics[width=0.3\textwidth ,valign=c]{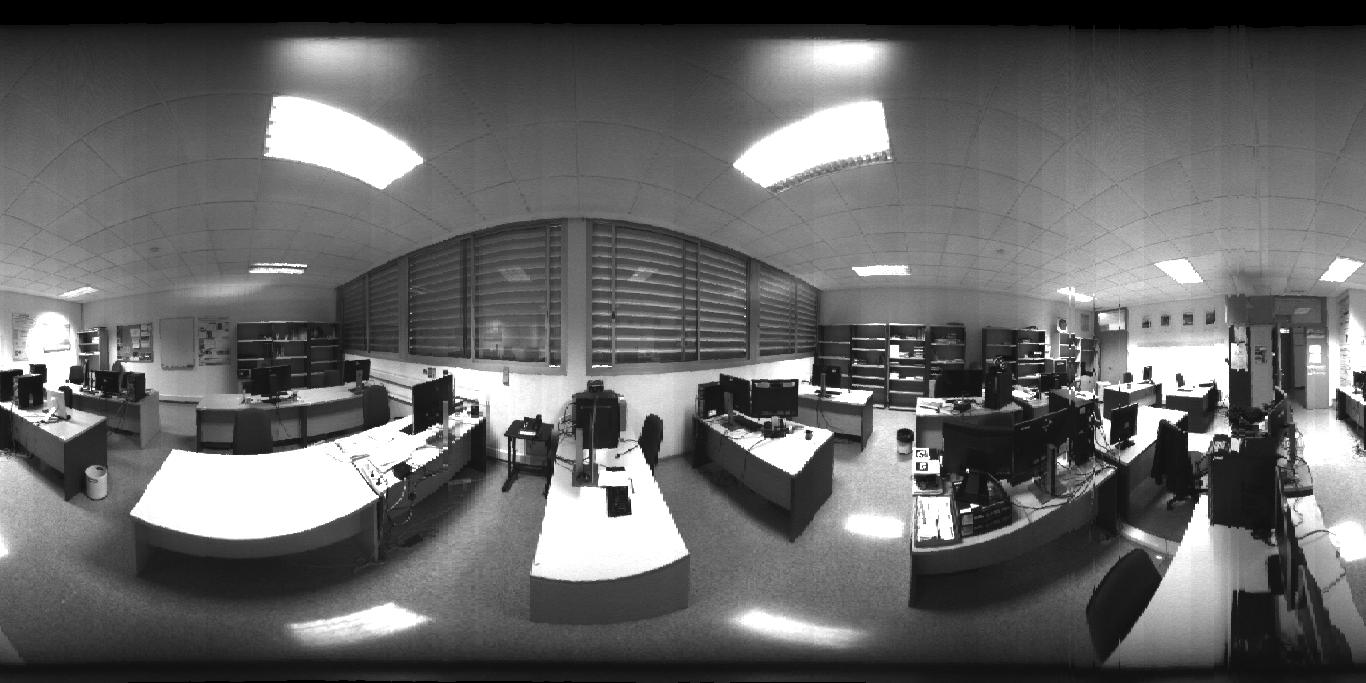}} \\
	\setcounter{subfigure}{0}
	\subfloat[\label{fig:realA}]{\includegraphics[width=0.33\textwidth ,valign=c]{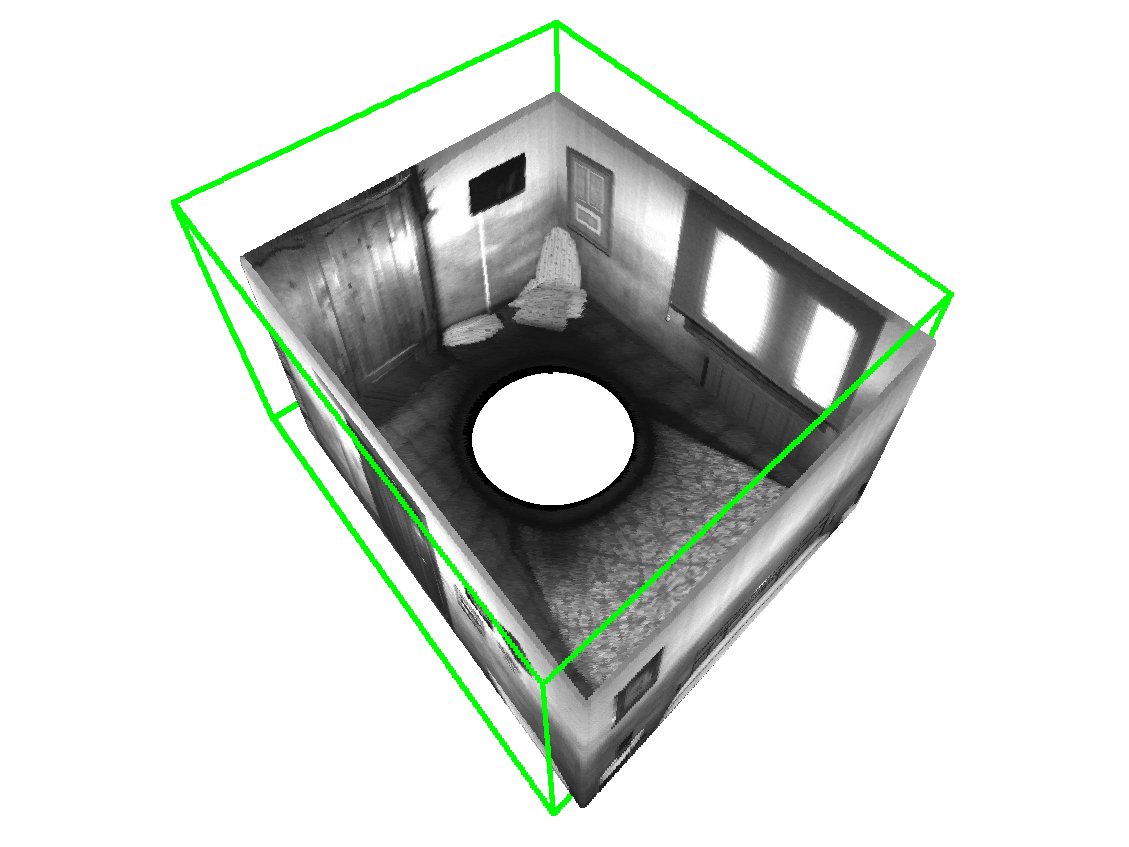}} \hfil
	\subfloat[\label{fig:realB}]{\includegraphics[width=0.33\textwidth ,valign=c]{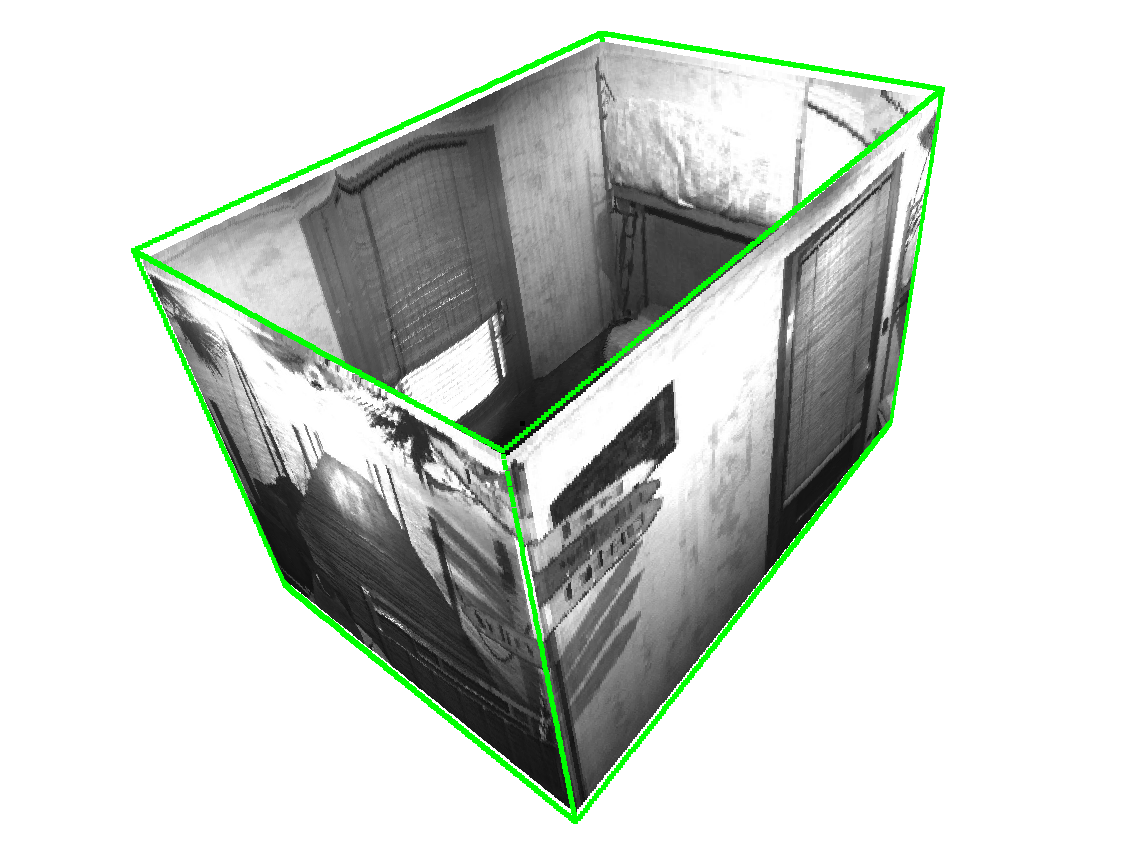}} \hfil
	\subfloat[\label{fig:realC}]{\includegraphics[width=0.33\textwidth ,valign=c]{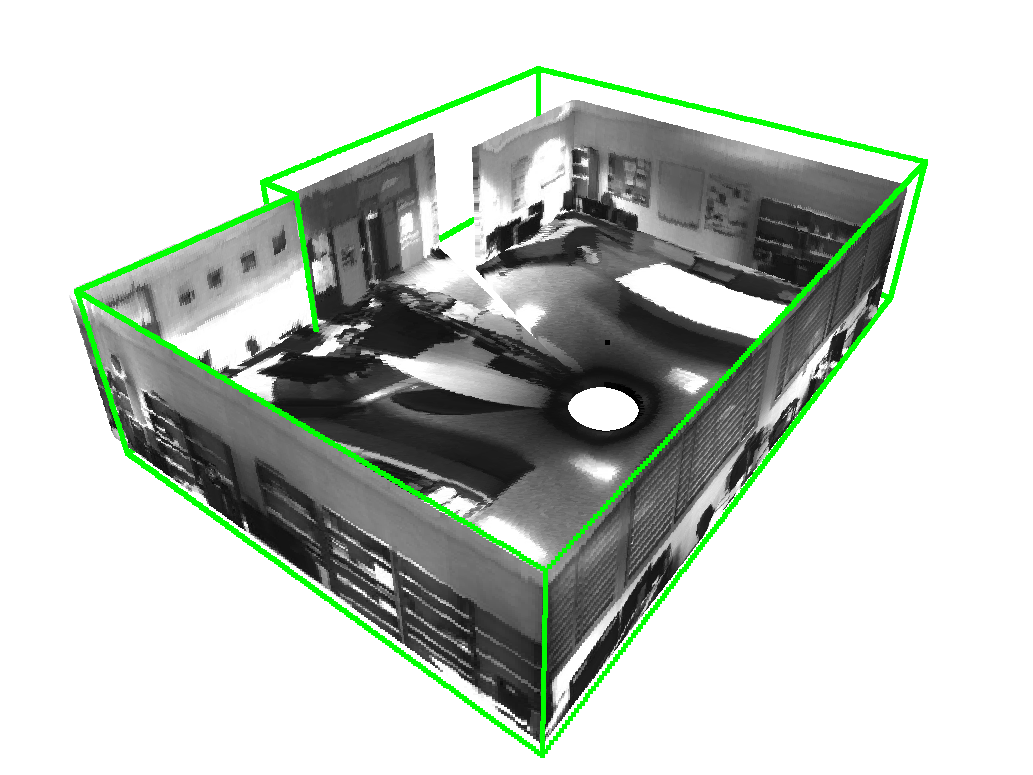}} 
	
	\caption{Examples of scaled layout reconstruction with real images. (a) shows an Atlanta environment; (b) and (c) show Manhattan environments.
	As qualitative evaluation, in green is a wire-frame of the room layout measured by hand with a laser meter.}
	\label{fig:realExp}
\end{figure}

In this section we present different examples of real non-central panoramas and the scaled layout reconstruction with our pipeline. 
The acquisition and anotation of these real examples has been made by the authors.

Figure \ref{fig:realExp} shows three different examples of real non-central panoramas, with different calibrations and in different environments. Fig. \ref{fig:realA} shows a bedroom that at first sight looked as Manhattan. After the reconstruction, it revealed itself as Atlanta, which was confirmed by the posterior measurement. The camera calibration in the acquisition of this image, defined by the radius of the non-central system, is $R_c = 0.5968 m$. The computed metrics of this reconstruction are: CE $= 0.65m$, CEN $= 3.97\%$, 3DIoU $= 54.71 \%$. Fig. \ref{fig:realB} shows other bedroom. This one is a Manhattan environment with a camera calibration $R_c = 0.6655 m$. The metrics of this reconstruction are: CE$ = 0.073 m$, CEN $= 0.92\%$, 3DIoU $= 91.94\%$. In Fig. \ref{fig:realC}, we show a computer laboratory, which is much bigger that the previous bedrooms. The calibration of the camera is $R_c = 0.6232 m$, and the metrics: CE $= 0.531 m$, CEN $= 1.12\%$, 3DIoU $= 74.97\%$.

\section{Discussion}
\label{sec:discussion}

\begin{figure*}
	\centering
	\subfloat{\includegraphics[width=0.24\textwidth ,valign=c]{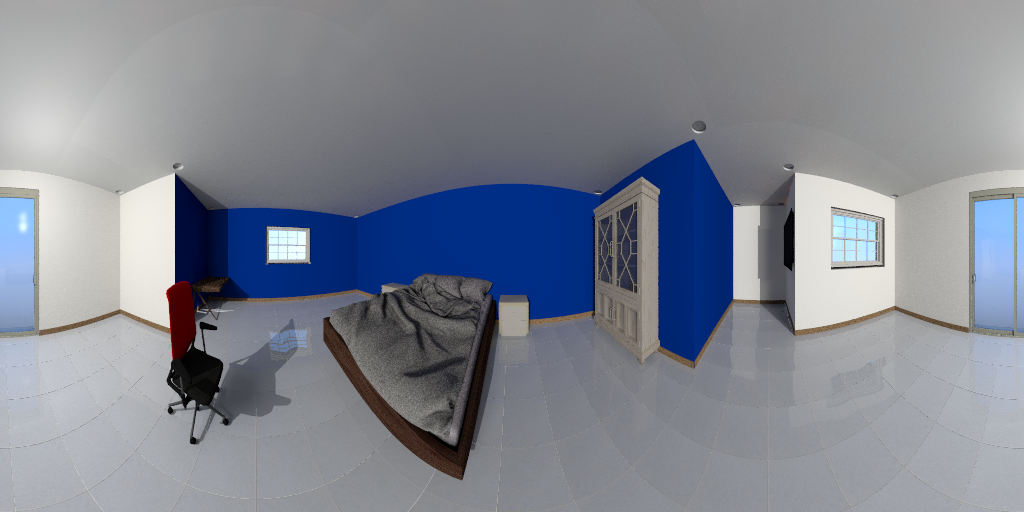}} \hfil
	\subfloat{\includegraphics[width=0.24\textwidth ,valign=c]{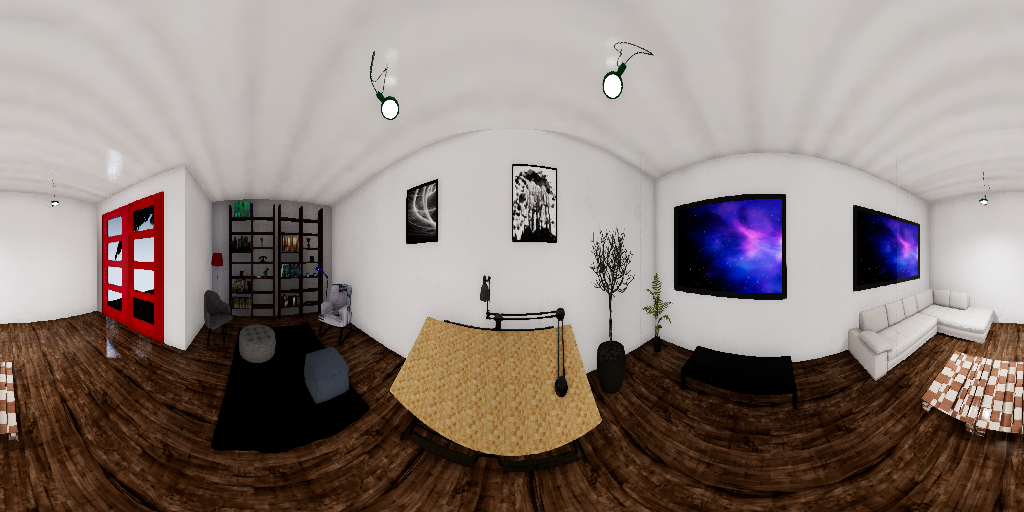}} 	\hfil
	\subfloat{\includegraphics[width=0.24\textwidth ,valign=c]{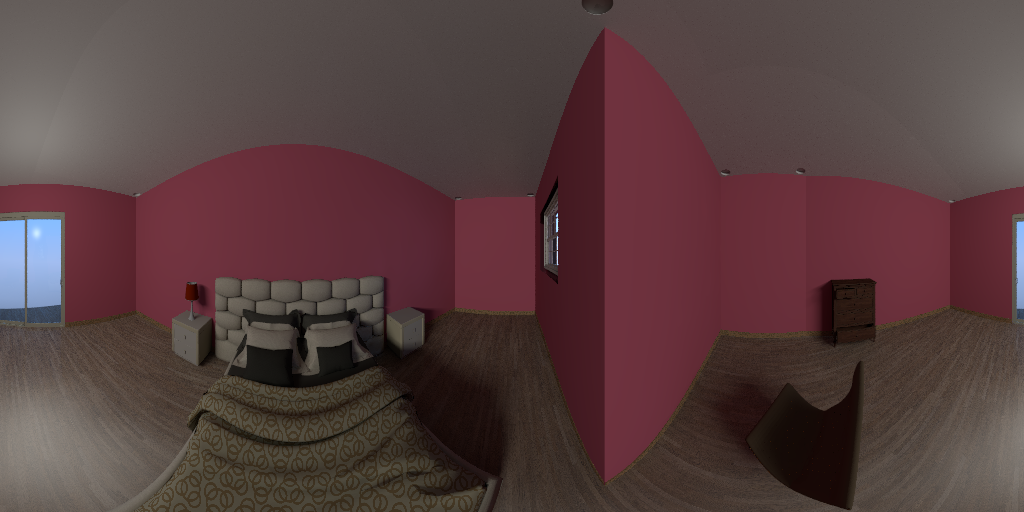}} 	\hfil
	\subfloat{\includegraphics[width=0.24\textwidth ,valign=c]{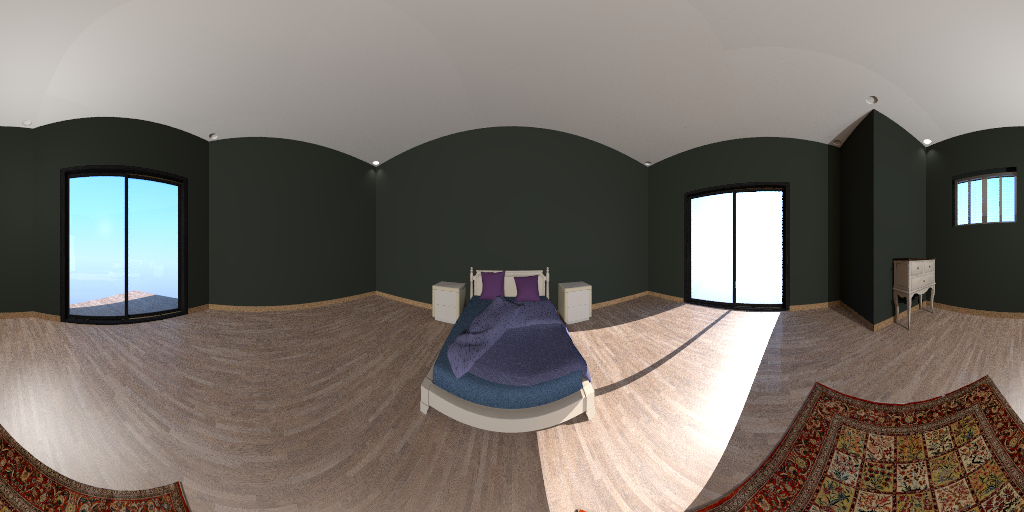}} \\
	
	\subfloat{\includegraphics[width=0.2\textwidth ,valign=c]{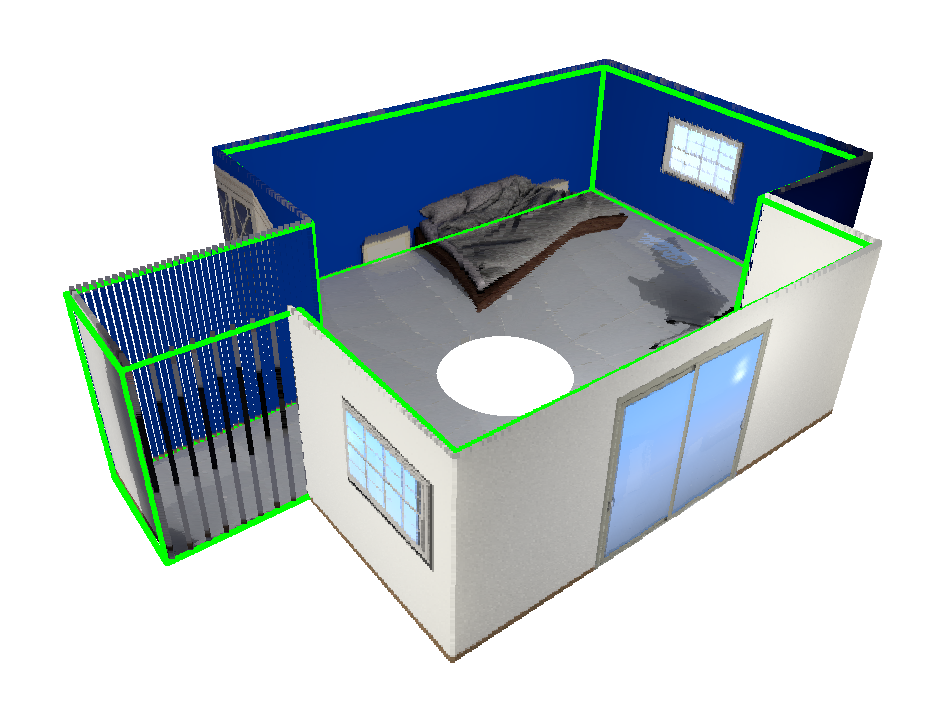}} \hfil
	\subfloat{\includegraphics[width=0.2\textwidth ,valign=c]{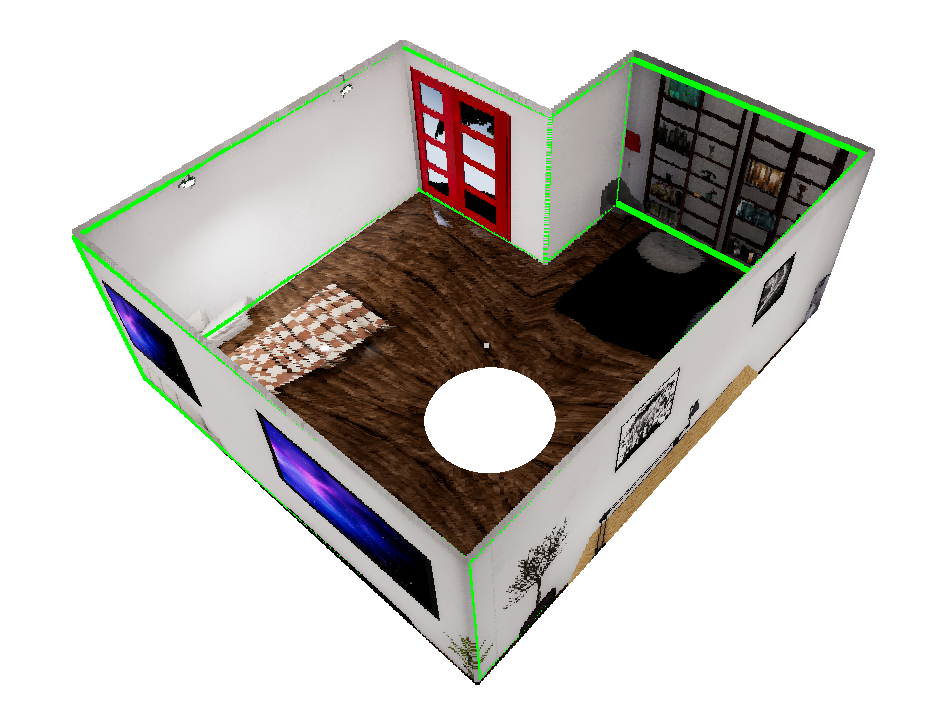}} 	\hfil
	\subfloat{\includegraphics[width=0.2\textwidth ,valign=c]{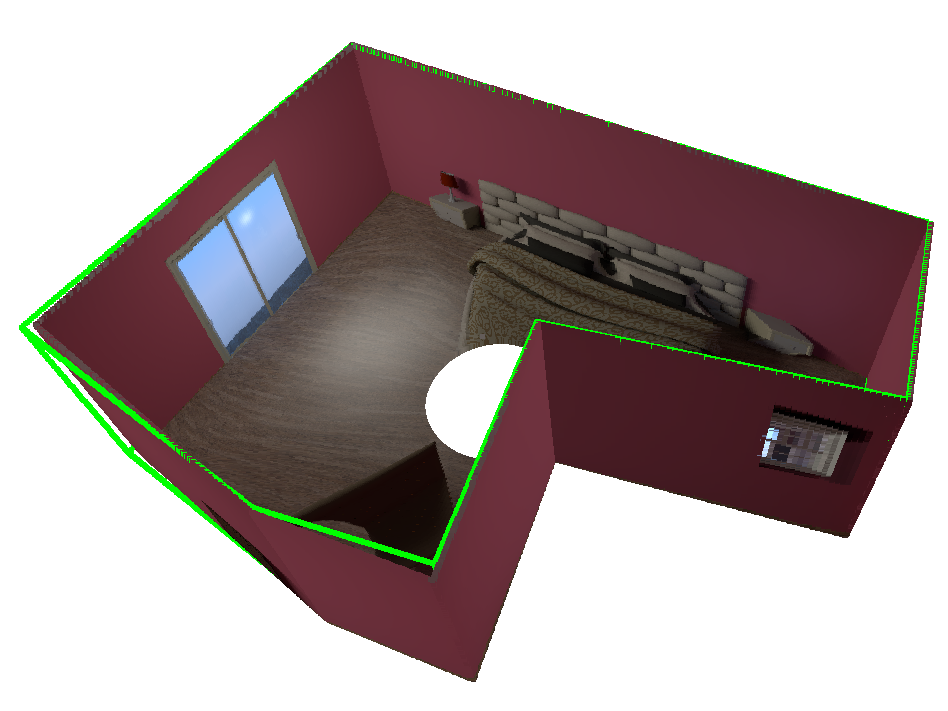}} 	\hfil
	\subfloat{\includegraphics[width=0.2\textwidth ,valign=c]{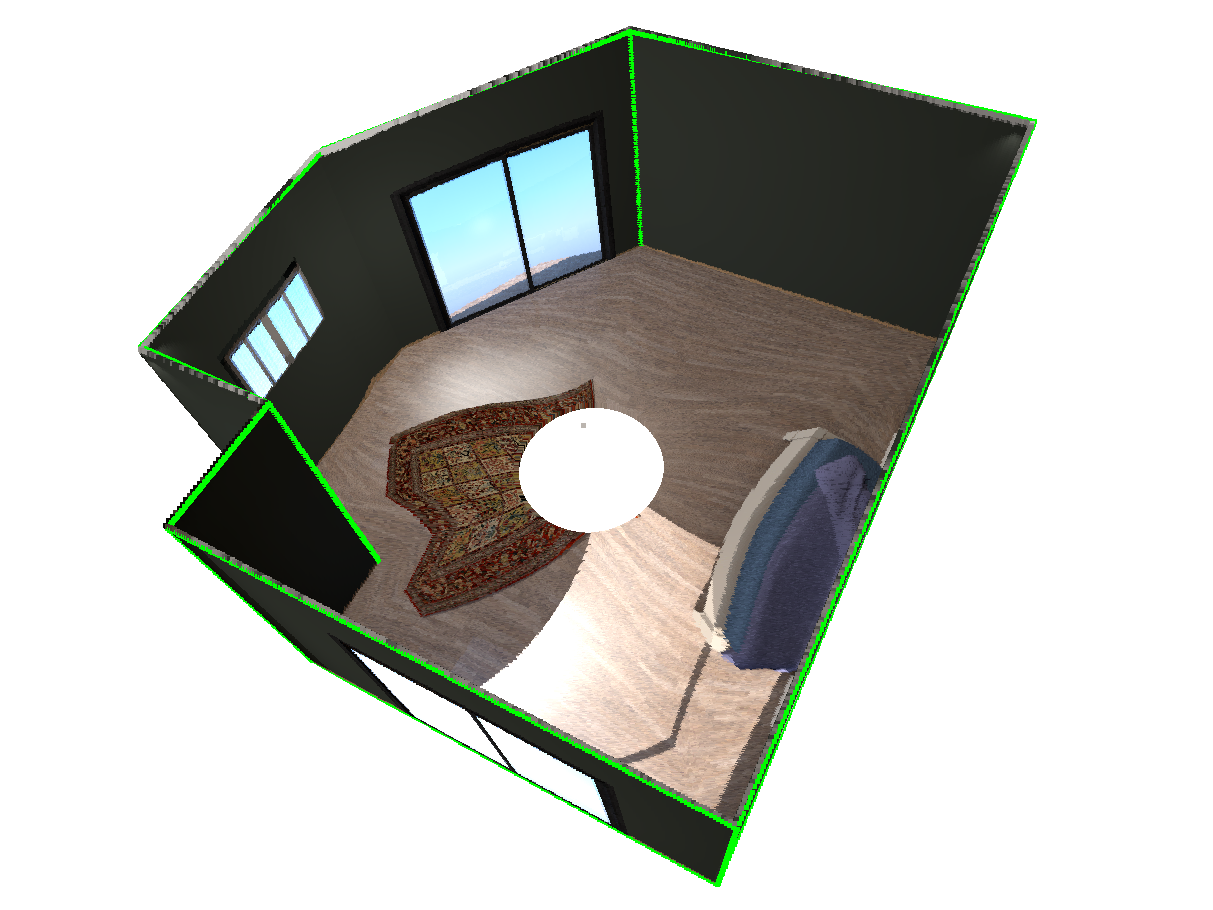}} \\
	\vspace{0.1cm}
	
	\subfloat{\includegraphics[width=0.24\textwidth ,valign=c]{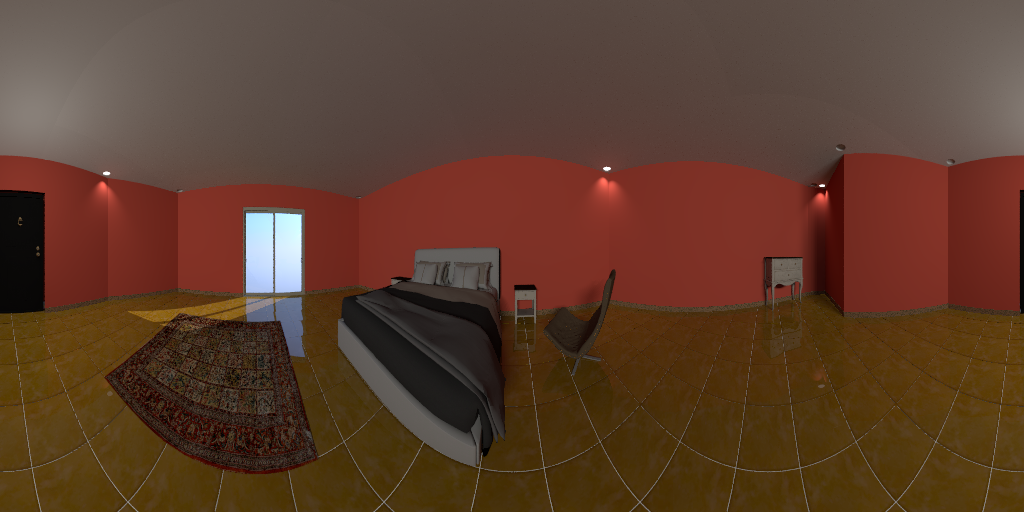}} \hfil
	\subfloat{\includegraphics[width=0.24\textwidth ,valign=c]{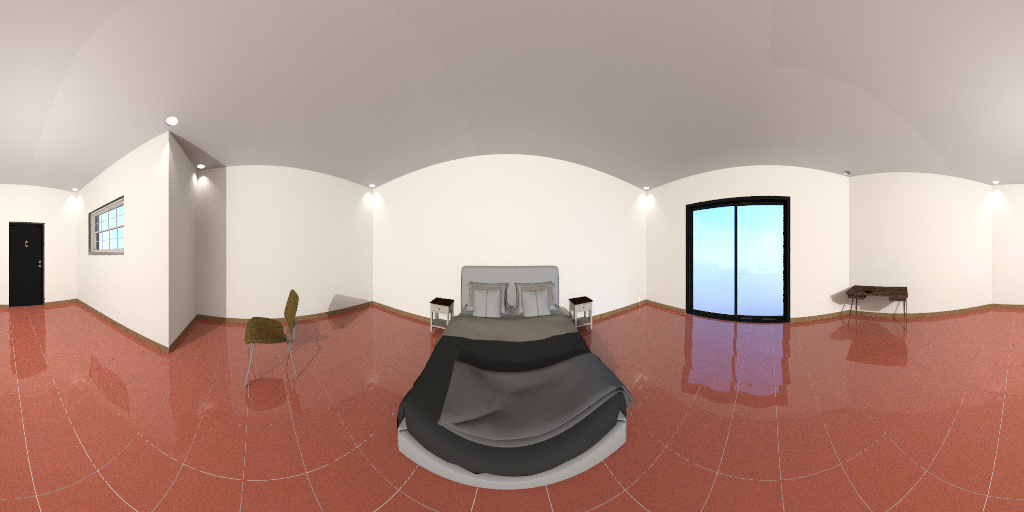}} \hfil
	\subfloat{\includegraphics[width=0.24\textwidth ,valign=c]{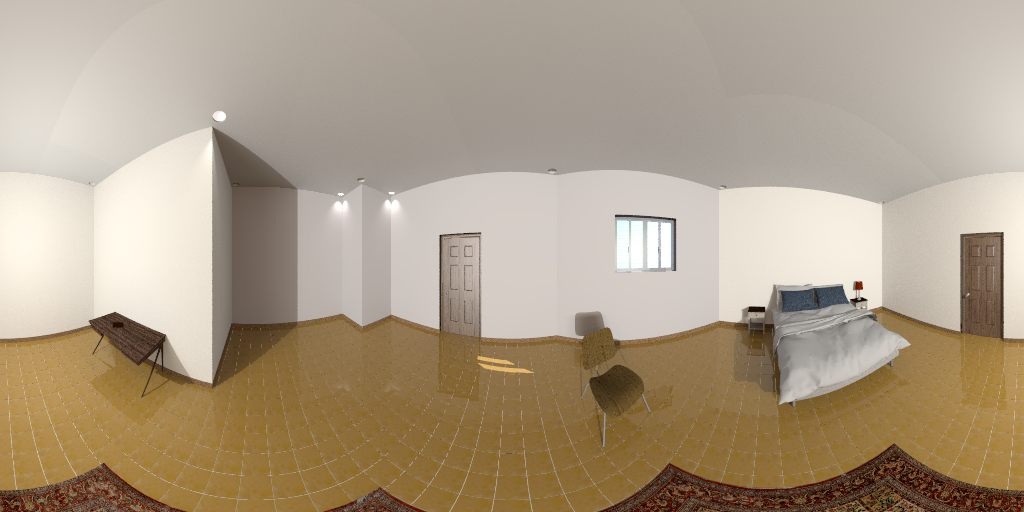}} \hfil
	\subfloat{\includegraphics[width=0.24\textwidth ,valign=c]{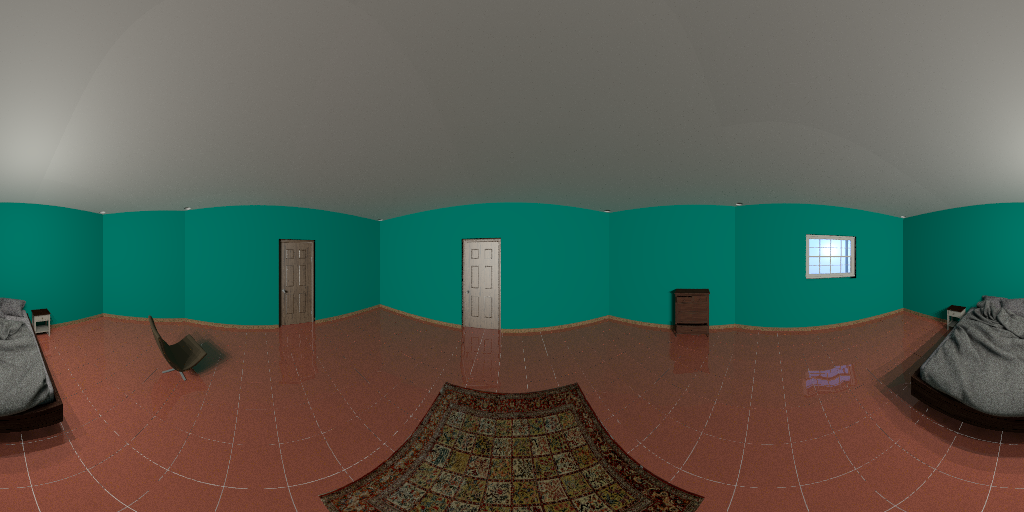}} \\
	
	\subfloat{\includegraphics[width=0.2\textwidth ,valign=c]{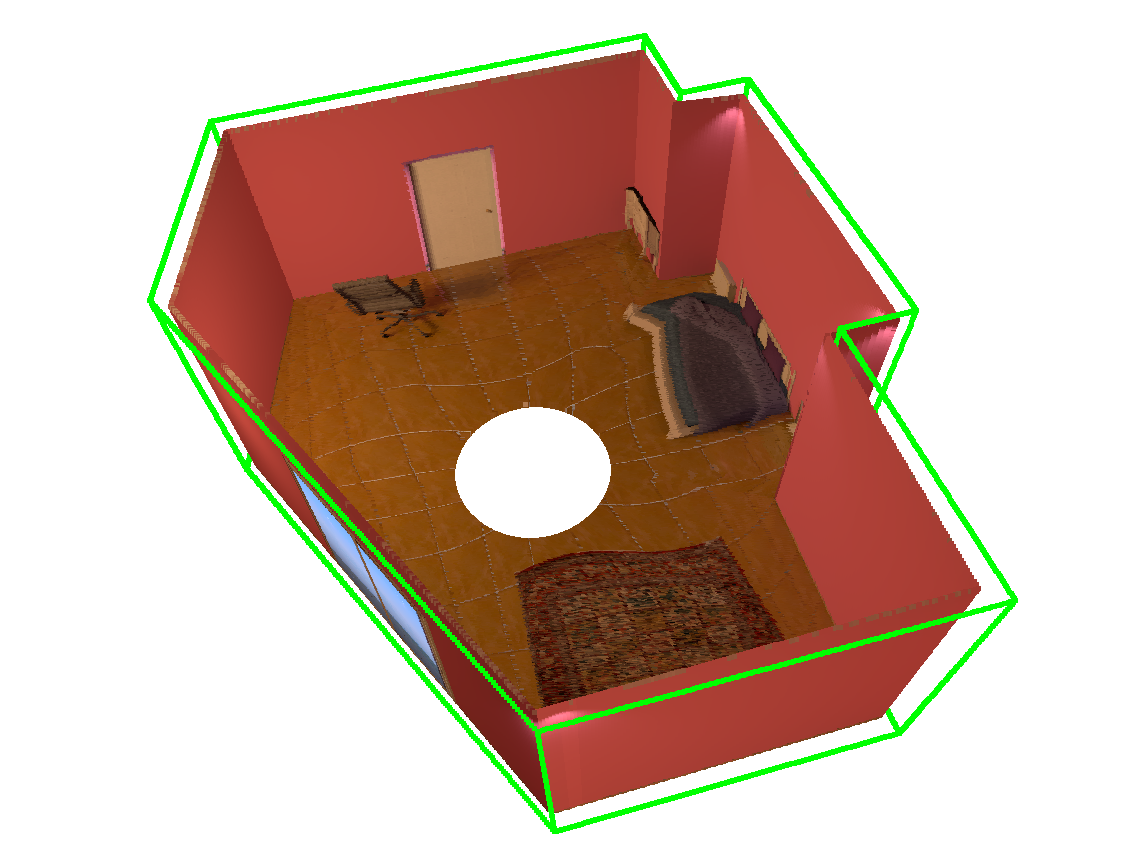}} \hfil
	\subfloat{\includegraphics[width=0.2\textwidth ,valign=c]{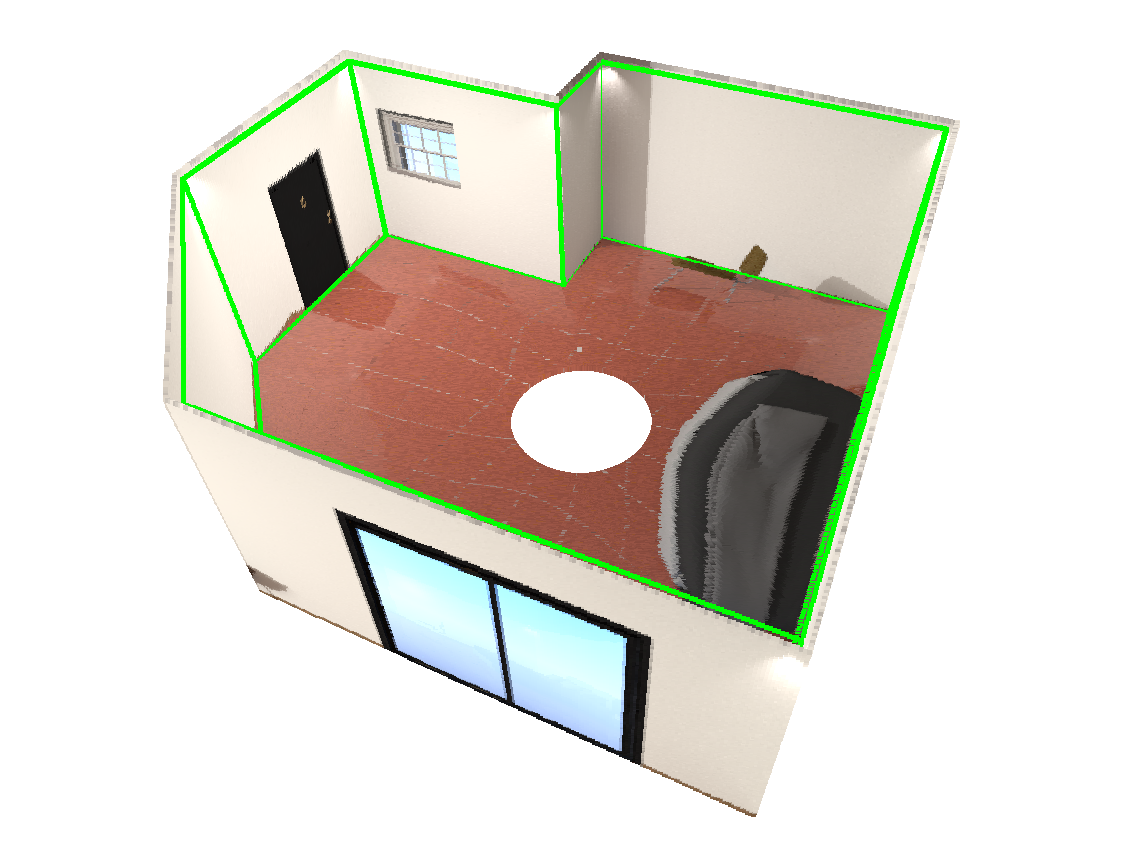}} \hfil
	\subfloat{\includegraphics[width=0.2\textwidth ,valign=c]{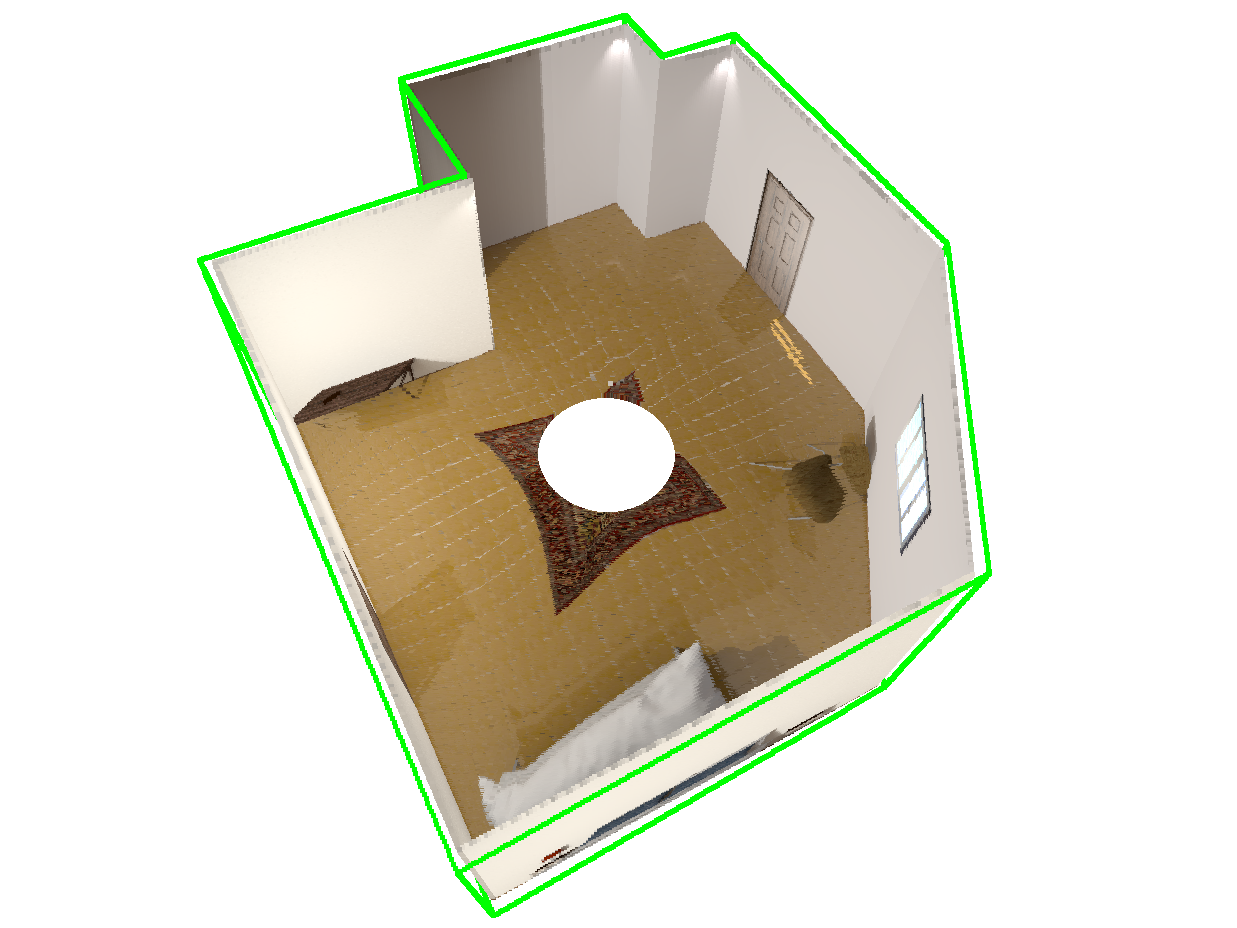}} \hfil
	\subfloat{\includegraphics[width=0.2\textwidth ,valign=c]{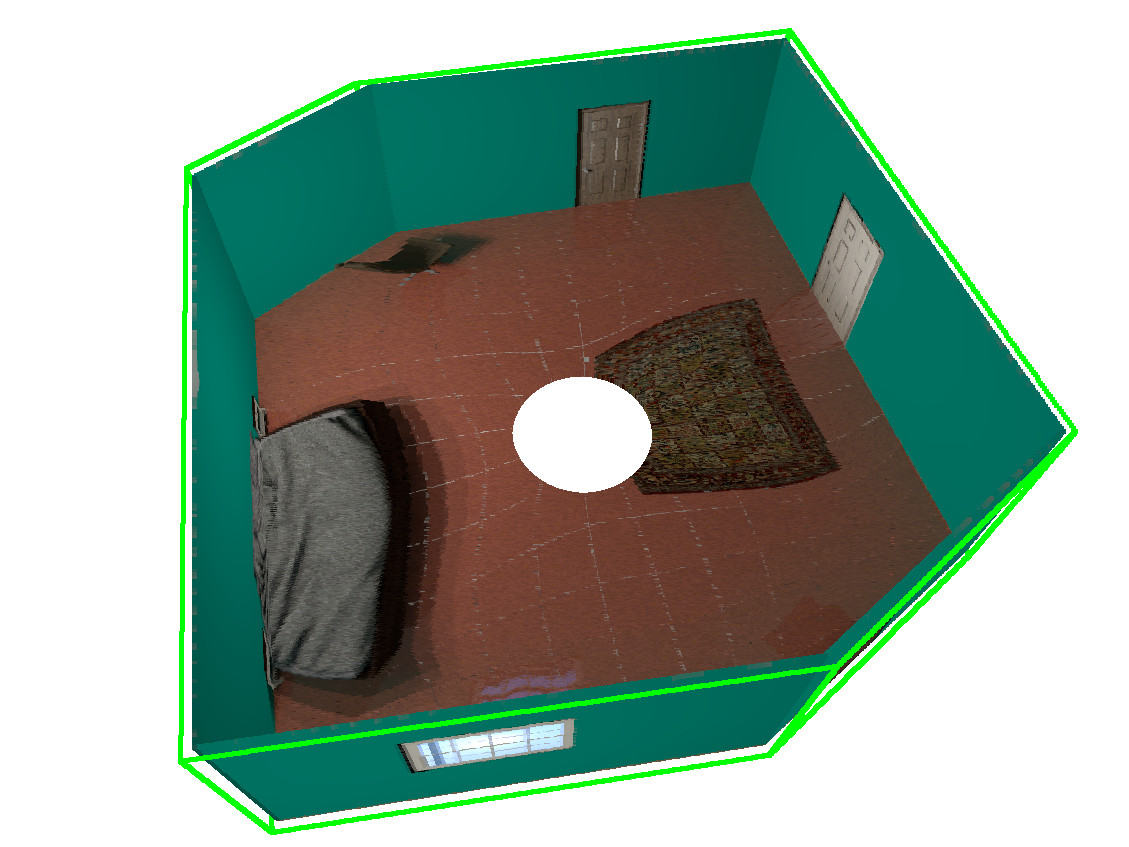}} 
	\vspace{0.1cm}
	
	\subfloat{\includegraphics[width=0.24\textwidth ,valign=c]{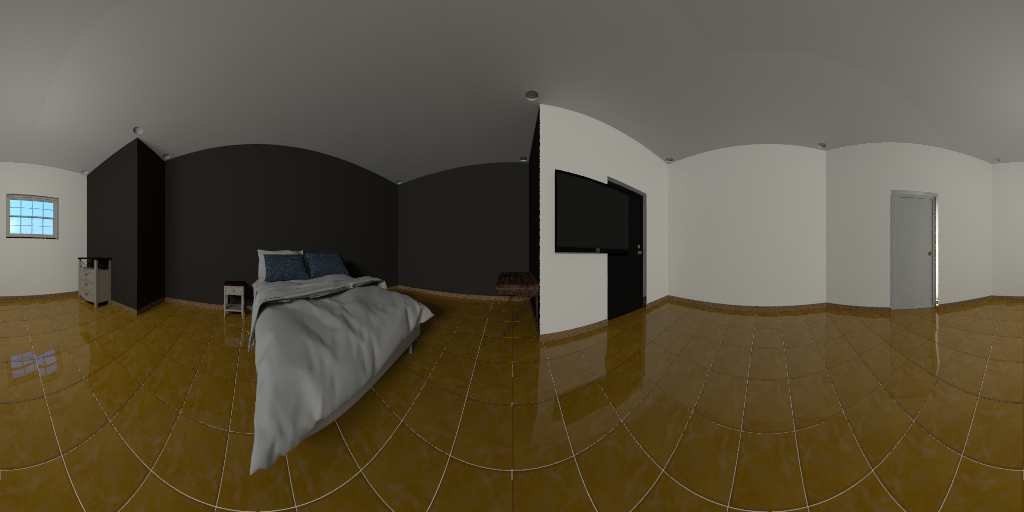}} \hfil
	\subfloat{\includegraphics[width=0.24\textwidth ,valign=c]{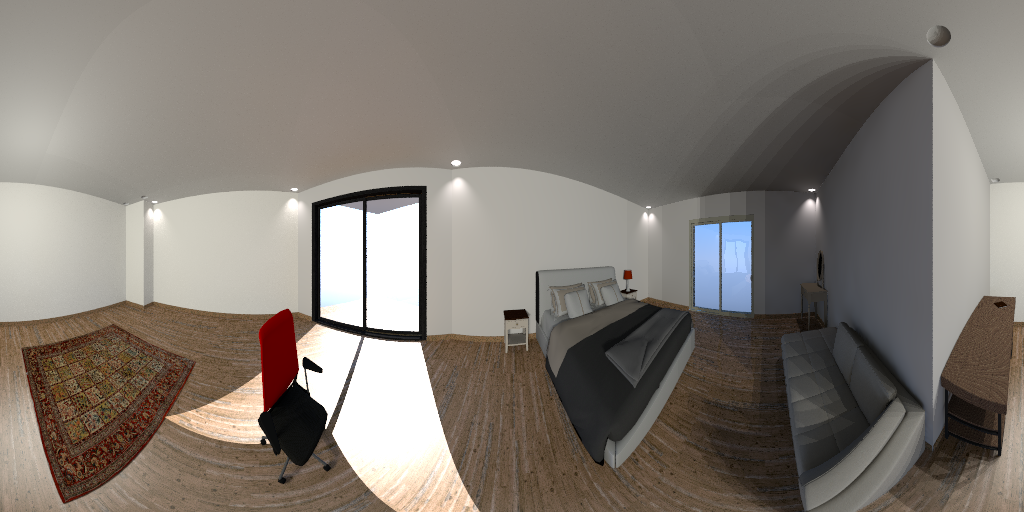}} 	\hfil
	\subfloat{\includegraphics[width=0.24\textwidth ,valign=c]{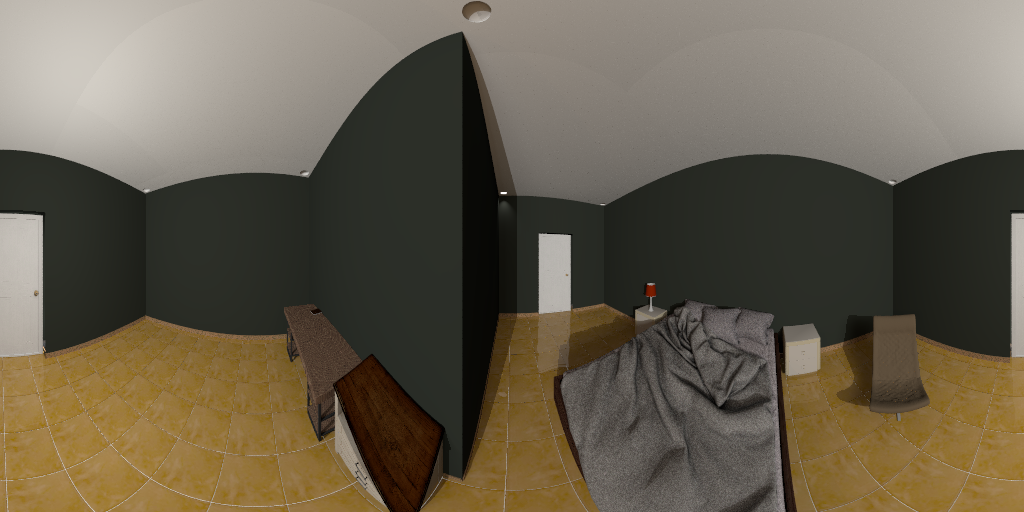}} 	\hfil
	\subfloat{\includegraphics[width=0.24\textwidth ,valign=c]{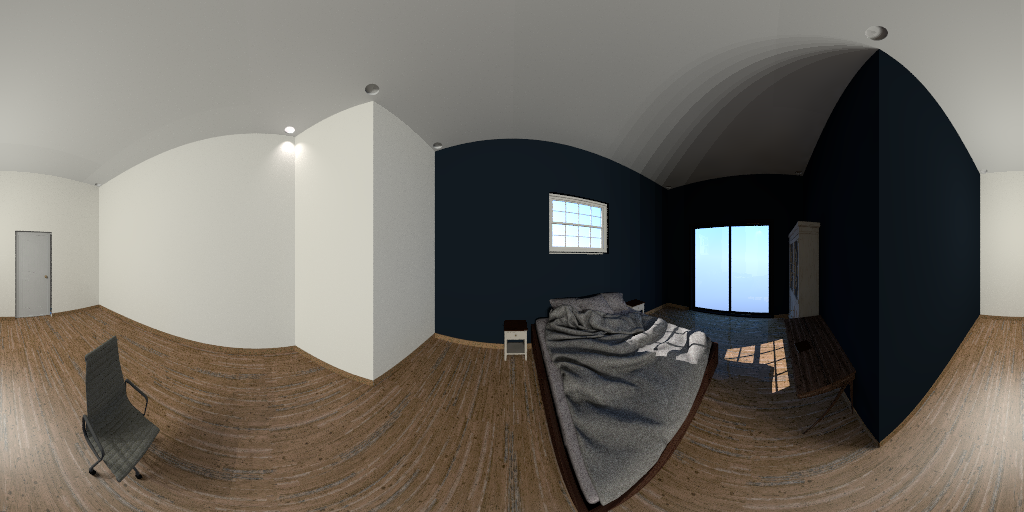}} \\
	
	\subfloat{\includegraphics[width=0.2\textwidth ,valign=c]{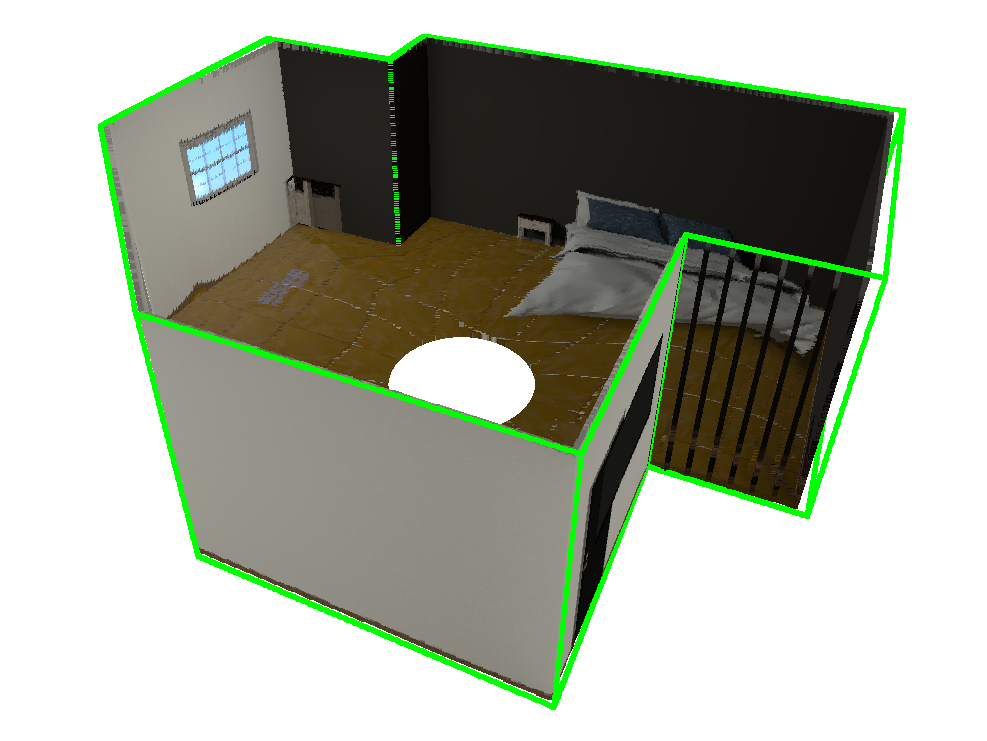}} \hfil
	\subfloat{\includegraphics[width=0.2\textwidth ,valign=c]{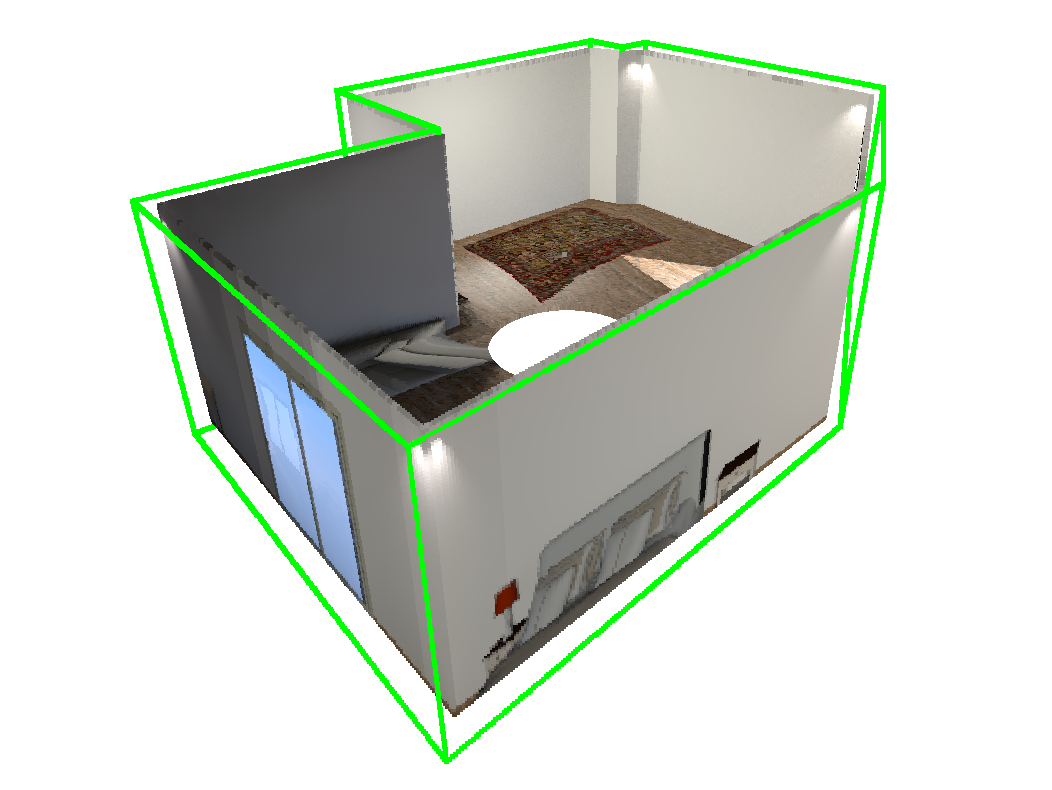}} 	\hfil
	\subfloat{\includegraphics[width=0.2\textwidth ,valign=c]{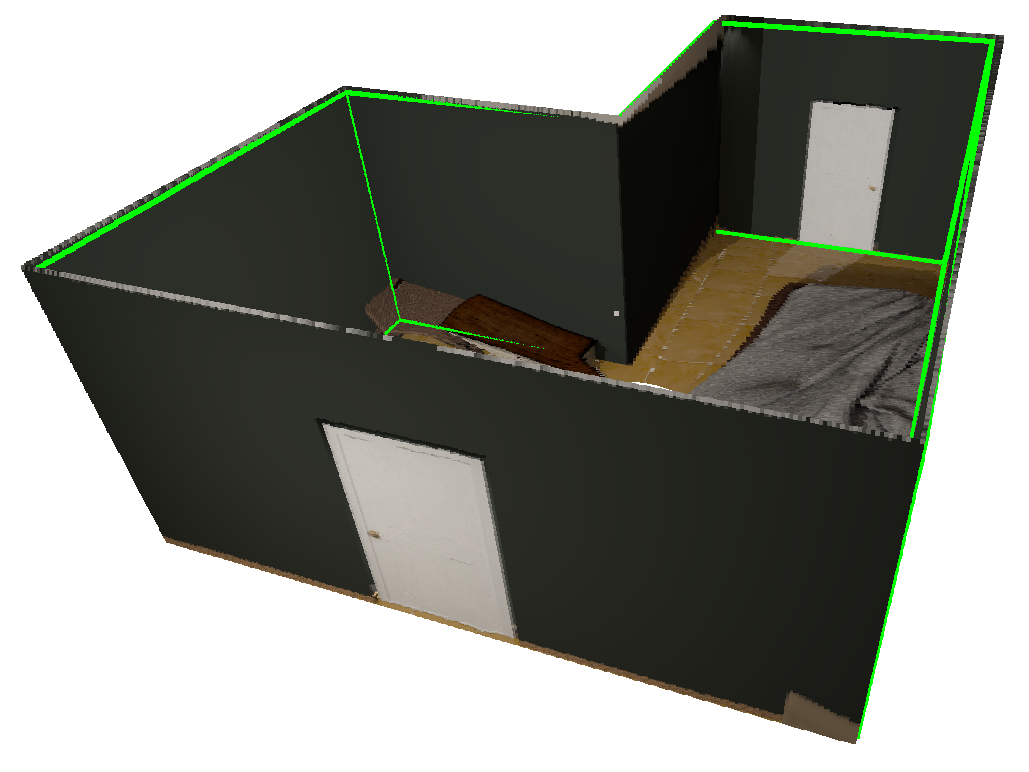}} 	\hfil
	\subfloat{\includegraphics[width=0.2\textwidth ,valign=c]{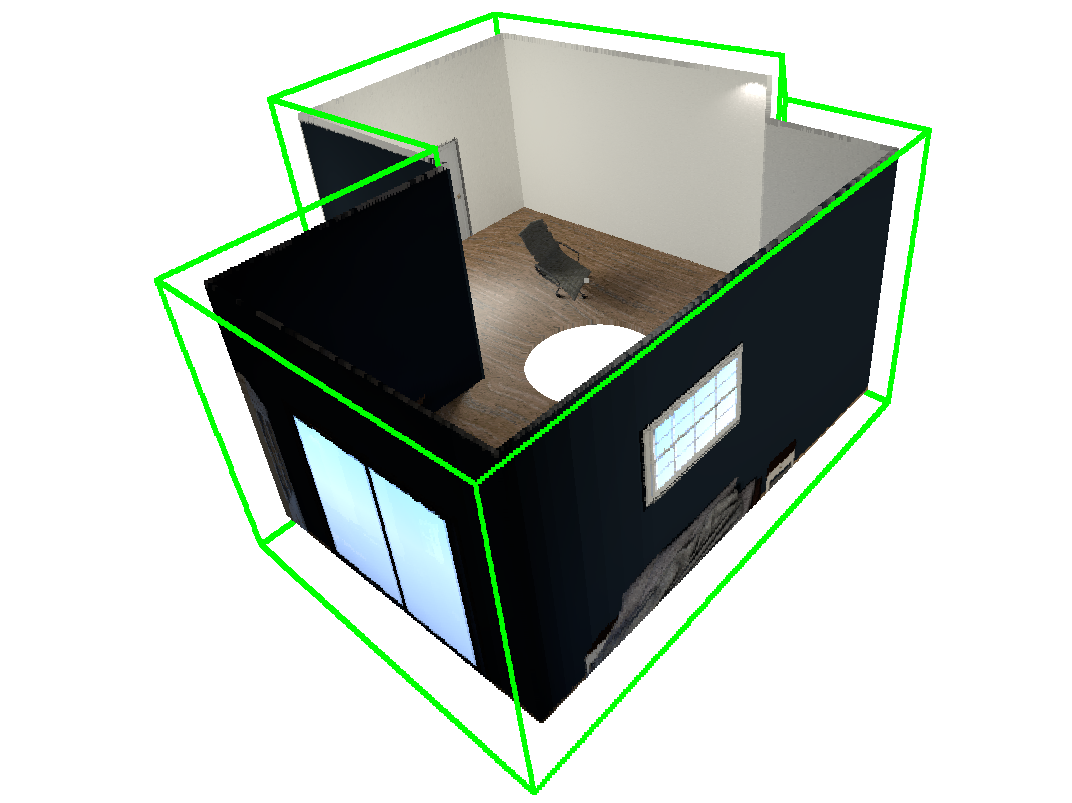}} \\
	\vspace{0.1cm}
	
	\subfloat{\includegraphics[width=0.24\textwidth ,valign=c]{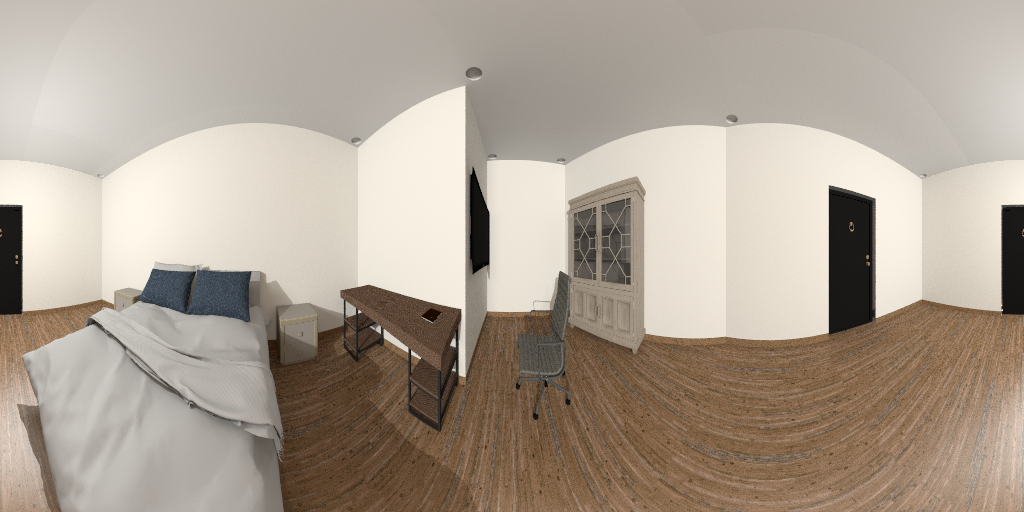}} \hfil
	\subfloat{\includegraphics[width=0.24\textwidth ,valign=c]{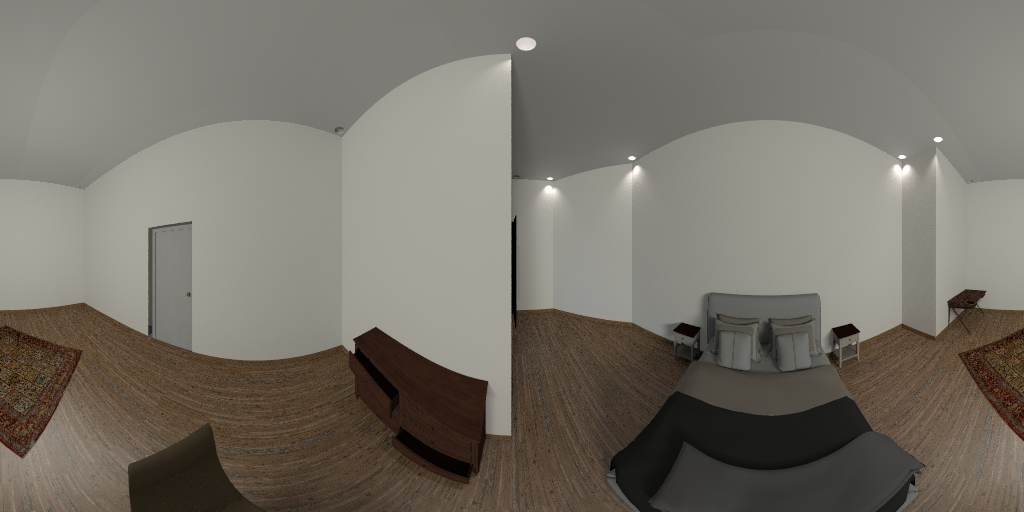}} \hfil
	\subfloat{\includegraphics[width=0.24\textwidth ,valign=c]{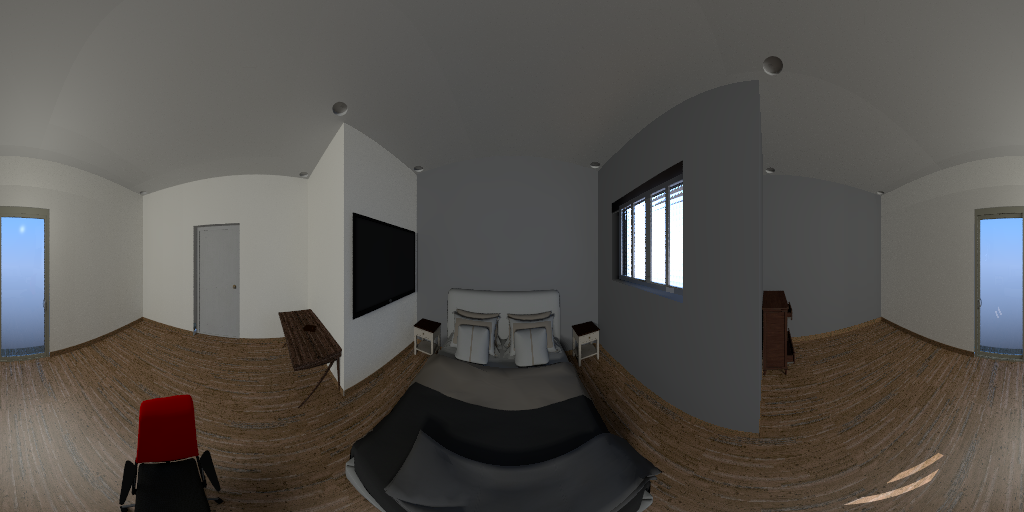}} \hfil
	\subfloat{\includegraphics[width=0.24\textwidth ,valign=c]{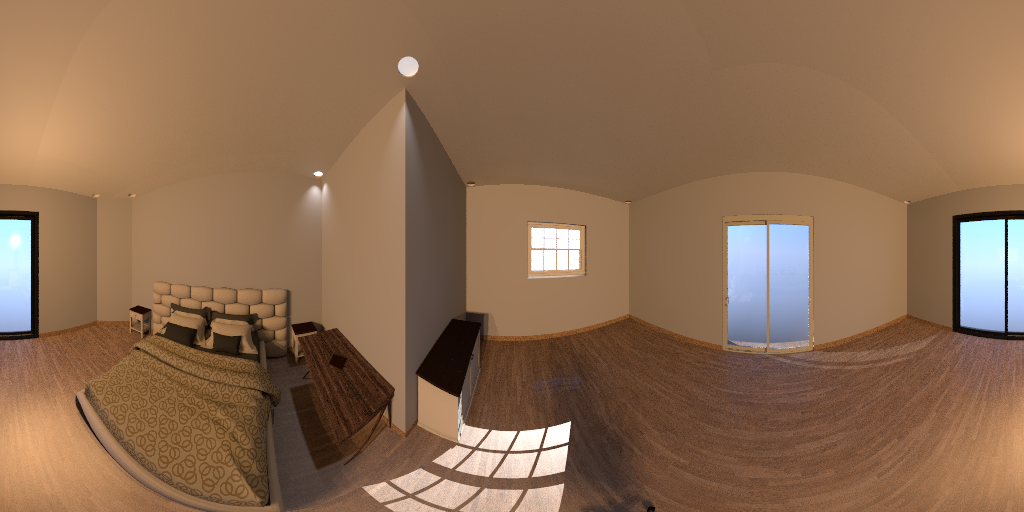}} \\
	
	\subfloat{\includegraphics[width=0.2\textwidth ,valign=c]{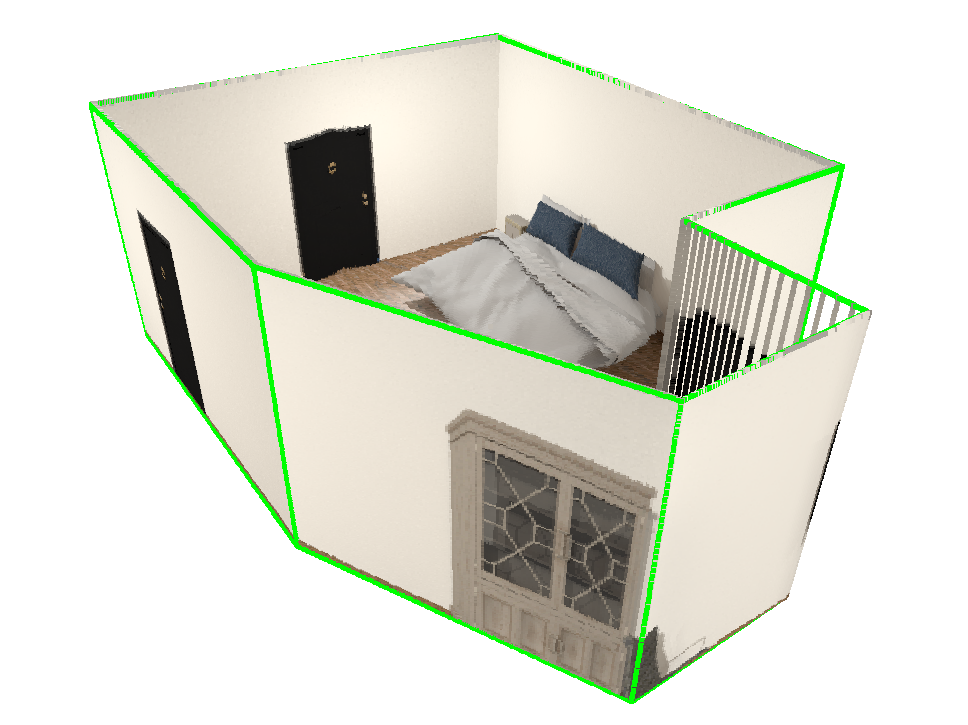}} \hfil
	\subfloat{\includegraphics[width=0.2\textwidth ,valign=c]{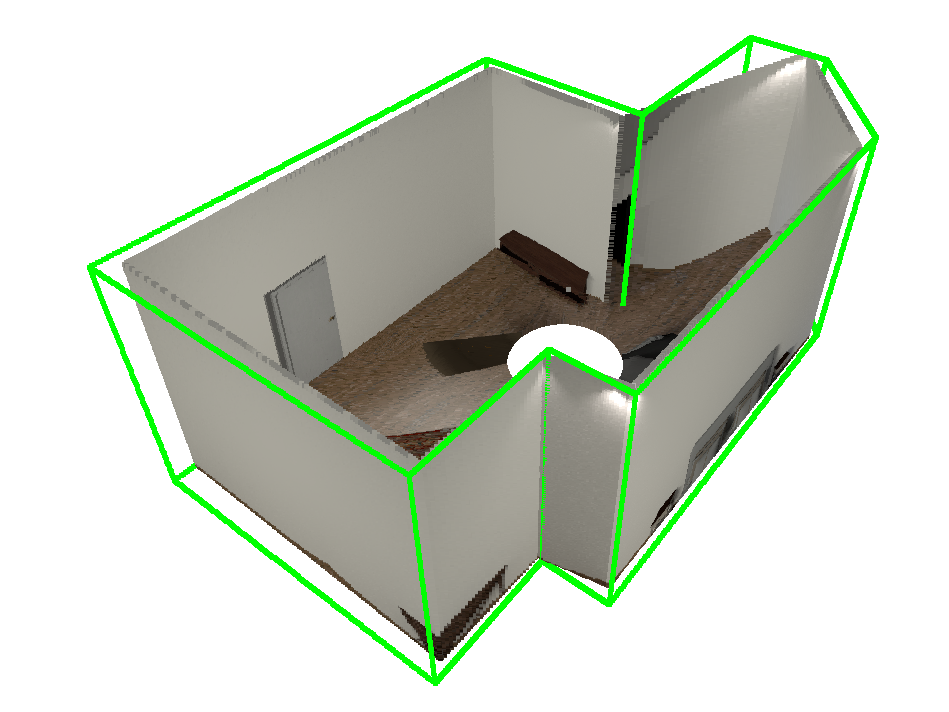}} \hfil
	\subfloat{\includegraphics[width=0.2\textwidth ,valign=c]{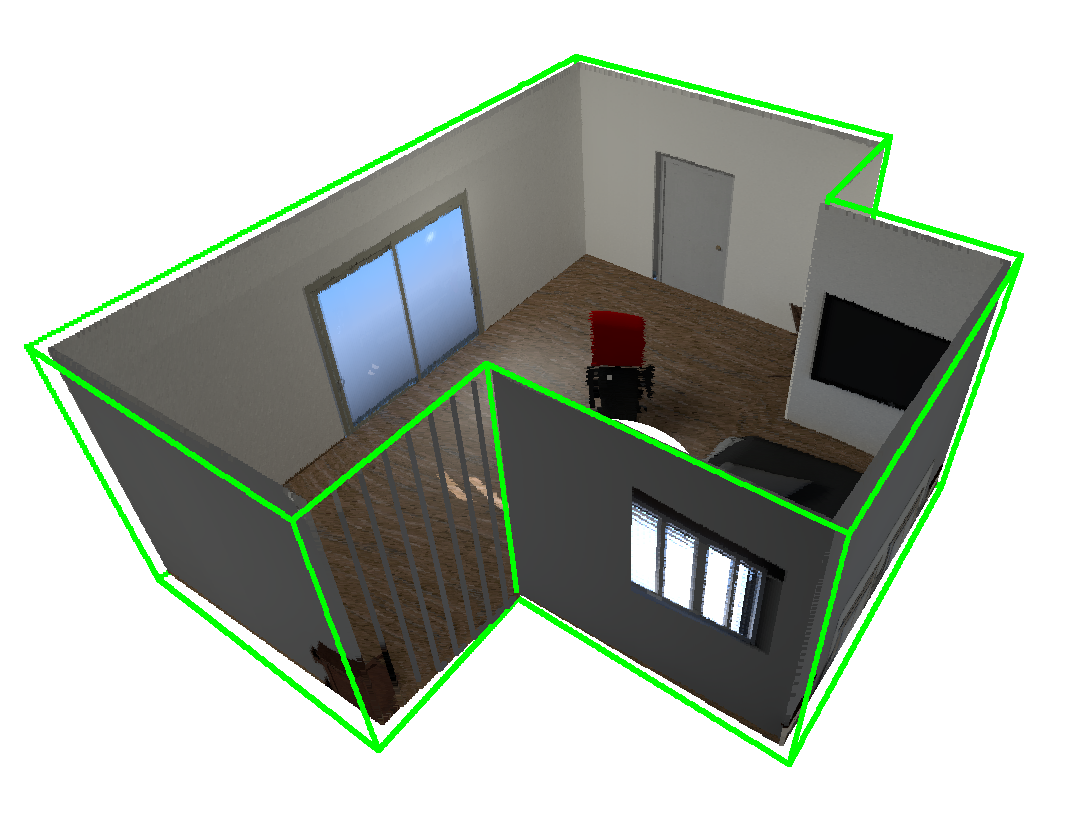}} \hfil
	\subfloat{\includegraphics[width=0.2\textwidth ,valign=c]{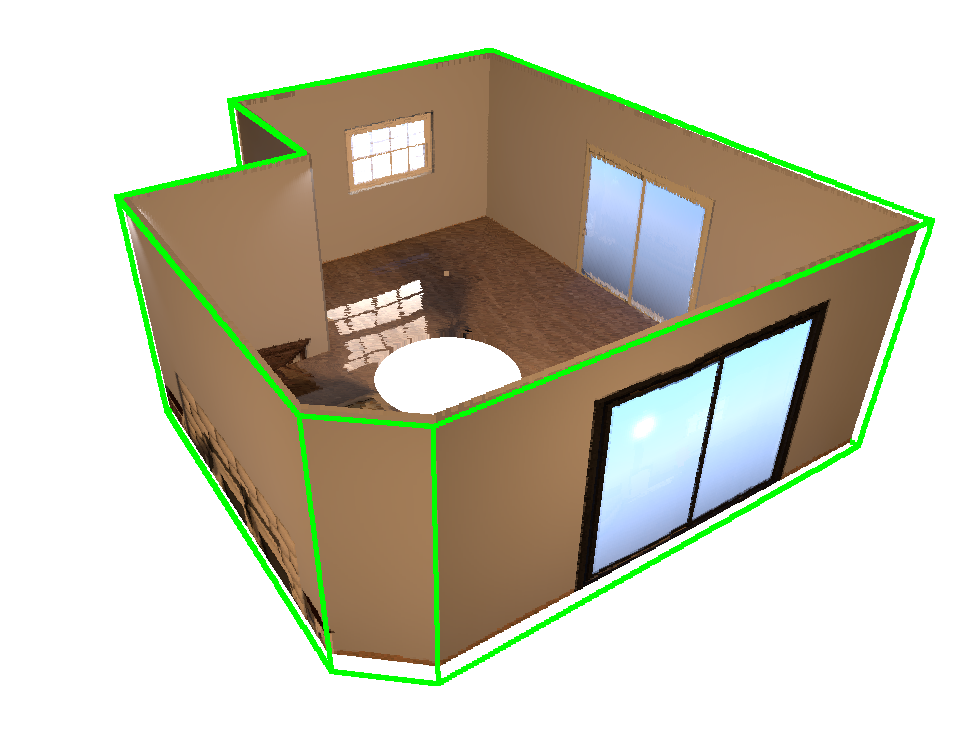}} 
	\caption{We present examples of non-central panoramas and its layout reconstruction from our method in different virtual environments of our data-set. The green wire-frame defines the ground truth layout of the environment.}
	\label{fig:results}
\end{figure*}

In this work we have presented a new method for layout recovery from a single non-central panorama. We have made a set of experiments to evaluate our method and compare it with the state of the art in layout recovery. In this section we analyse and discuss the results obtained in the previous experiments. 

Sections \ref{subsec:network_exp}, \ref{subsec:geometry_exp} and \ref{subsec:pipe_exp} evaluate different parts of our pipeline. In the first one, we evaluate how the performance of the proposed neural network improves from the original implementation in \cite{sun2019horizonnet} to our fine-tuned model with non-central panoramas. From this experiment we infer that the central and non-central distortions in panoramas are similar but not equal for the neural network. With a fine-tune of the network, we can adapt it to non-central panoramas, which provide more information of the environment than equirectangular panoramas.

In \ref{subsec:geometry_exp} we compare our new layout solvers against a state-of-the-art line extractor for non-central panoramas. We observe that our approach outperforms previous methods for line extraction in both Manhattan and Atlanta world assumptions by a large margin.
Finally, the evaluation of our geometric block validates the use of a more complex geometric pipeline. From the results presented in Table \ref{tab:pipeExp}, we observe that the difference in performance between the solvers and the pipeline using the network labels in Manhattan environments is quite small. However, for Atlanta environments and with the use of the network predictions, the performance of the full geometric pipeline is significantly better.

In section \ref{subsec:comparison} we compare our method with different state of the art implementations for layout recovery. In a first experiment, we make a comparison with HorizonNet. This is the most fair comparison made since we use panoramas from the same virtual environments and same locations on both methods in order to recover their layout. The results show that our proposal outperforms HorizonNet in the two cases of study: with extra measurements for HorizonNet and in up-to-scale reconstructions. On the second experiment, we make a comparison with other state of the art methods. In this experiment, the datasets used for the experiments are different, so the results do not completly show the performance of each method. Nevertheless, making the comparison of the results of each method, our proposal presents a better performance in most of the metrics. Besides, we are able to recover the scale of the environment without extra measurements while other methods need a metric measure to scale the layout (e.g the camera or room height).

Furthermore, we want to highlight the results in Atlanta environments. State-of-the-art methods do not refer in their works the management of occlusions in Atlanta environments, although they do manage occlusion in Manhattan world. Our proposal does handle occlusions in Atlanta environments, as well as in Manhattan environments. The results shown in Table \ref{tab:comparison} for our proposal include Manhattan and Atlanta environments with occlusions, which we are able to handle.

\begin{figure*}
\centering
	\subfloat{\includegraphics[width=0.24\textwidth ,valign=c]{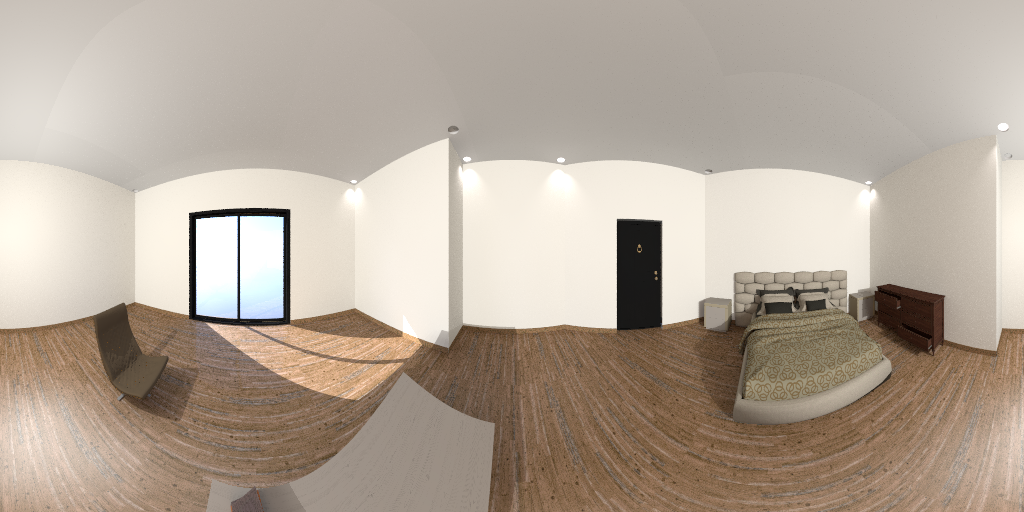}} \hfil
	\subfloat{\includegraphics[width=0.24\textwidth ,valign=c]{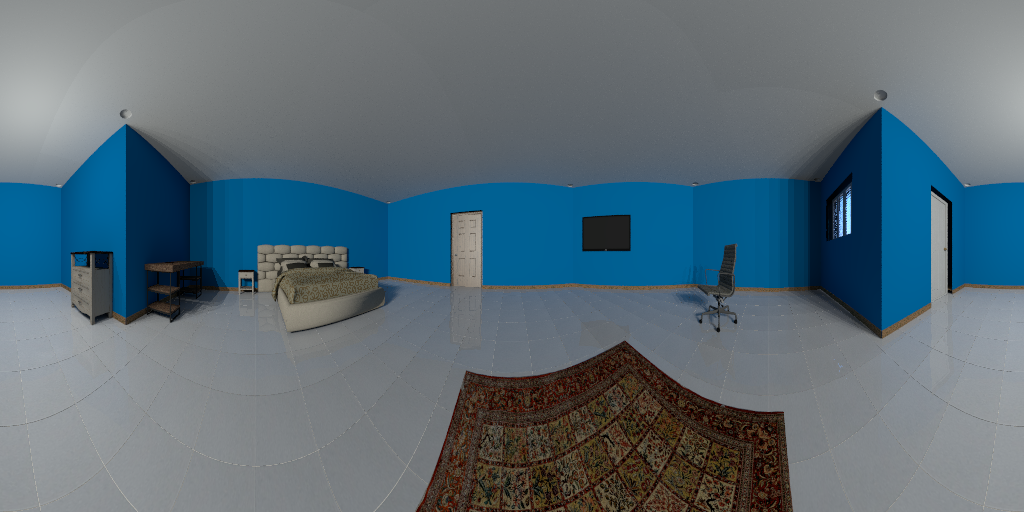}} \hfil
	\subfloat{\includegraphics[width=0.24\textwidth ,valign=c]{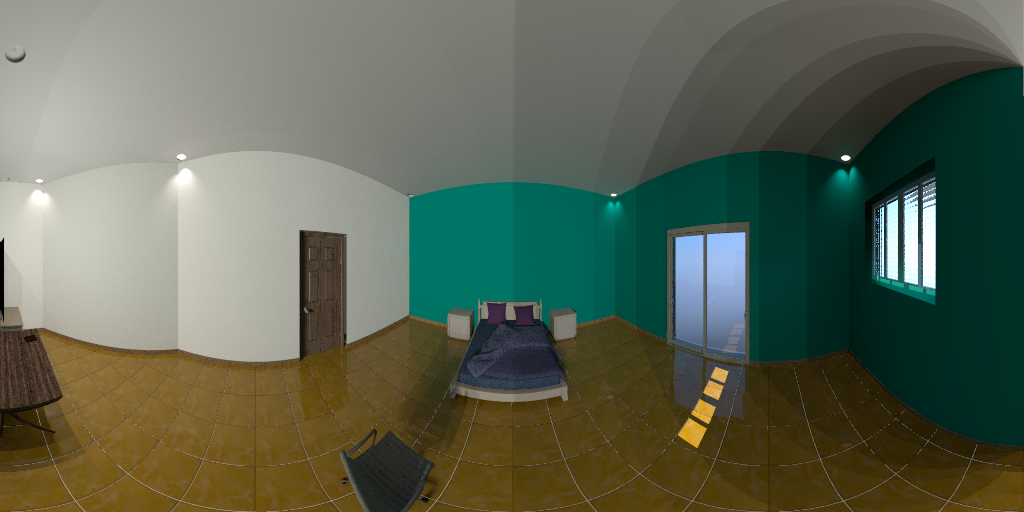}} \hfil
	\subfloat{\includegraphics[width=0.24\textwidth ,valign=c]{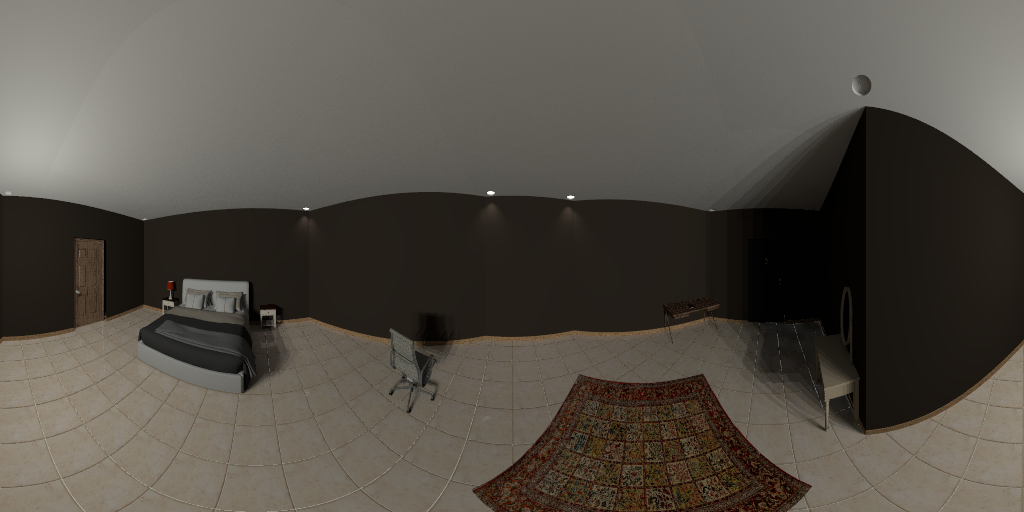}} \\
	\subfloat{\includegraphics[width=0.2\textwidth ,valign=c]{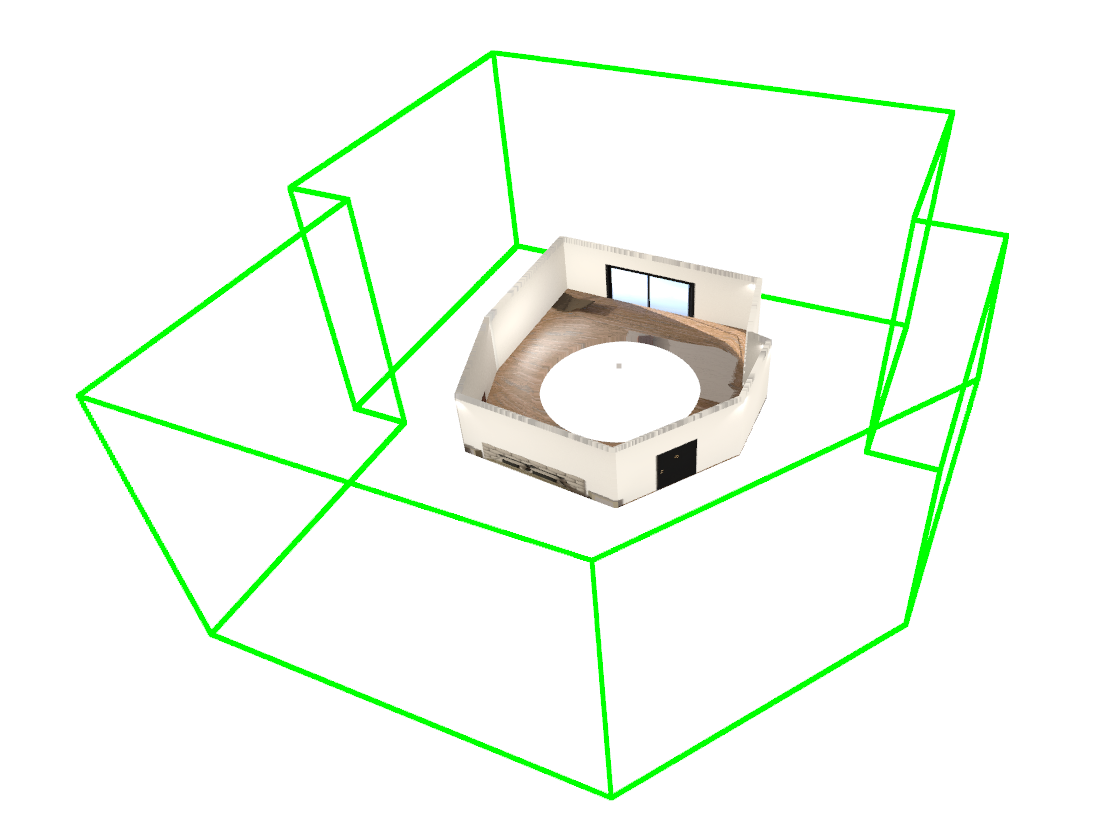}} \hfil
	\subfloat{\includegraphics[width=0.2\textwidth ,valign=c]{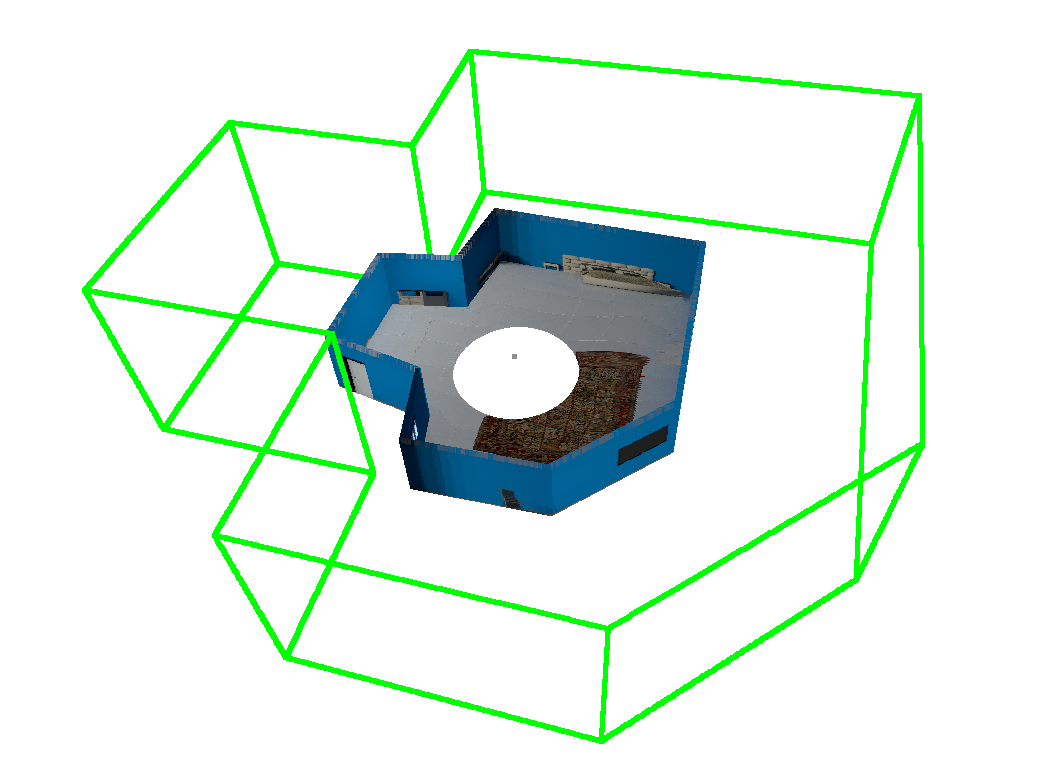}} \hfil
	\subfloat{\includegraphics[width=0.2\textwidth ,valign=c]{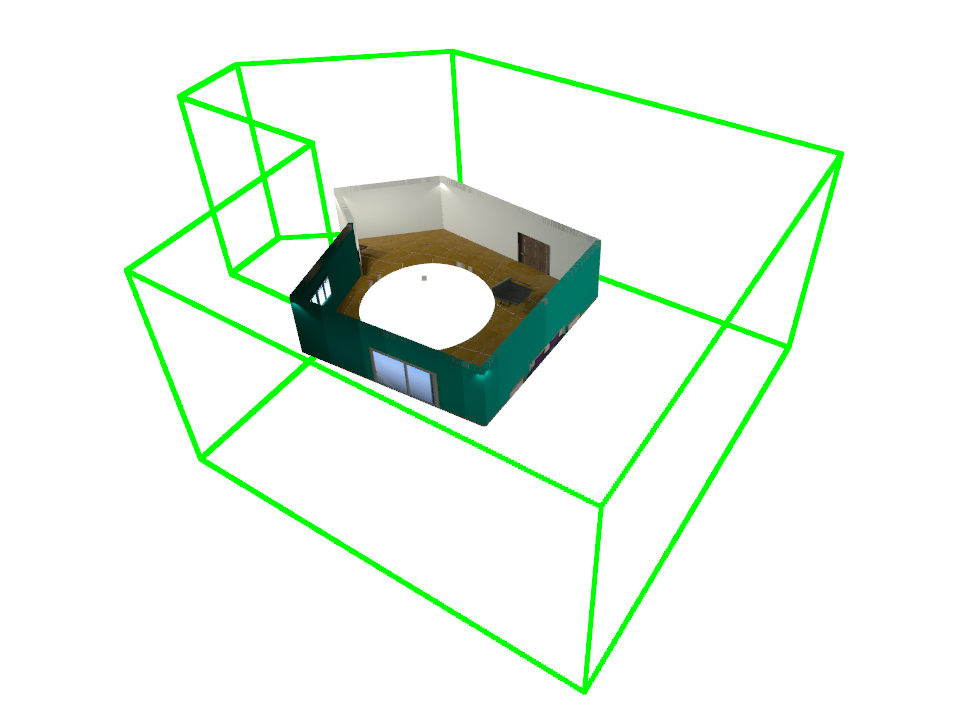}} \hfil
	\subfloat{\includegraphics[width=0.2\textwidth ,valign=c]{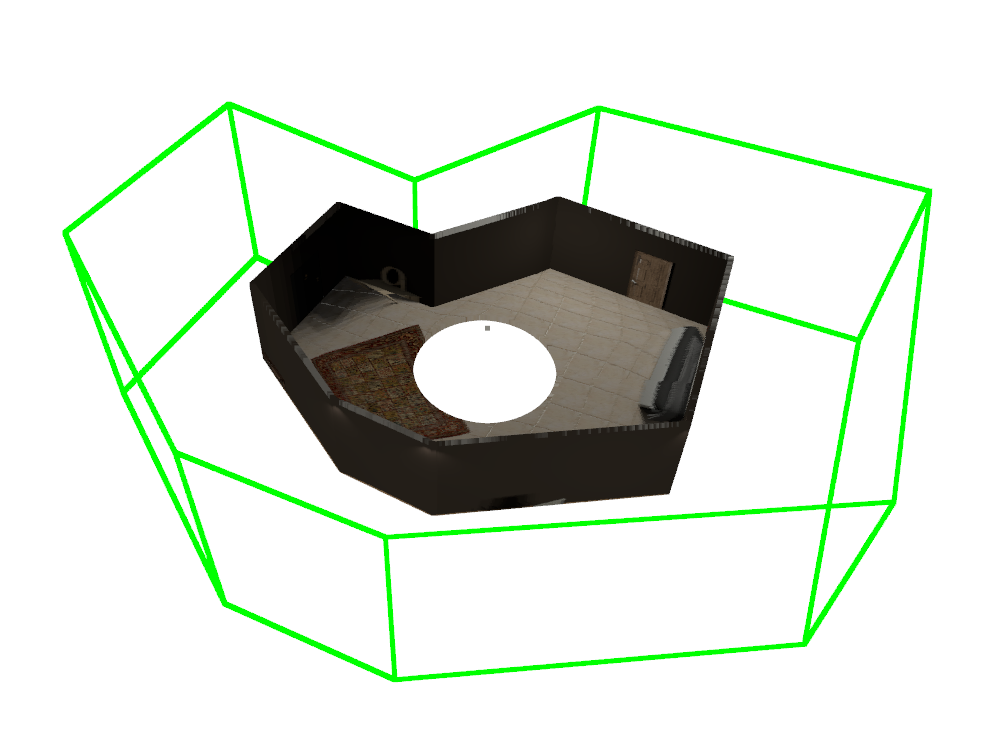}} 
	\caption{We present cases where our method fails in the scale recovery. The main sources of error are the output of the network, which may provide too noisy information; the occlusions in the Atlanta environment, which can misslead the final adjustment; or an insuficient efective baseline of the non-central acquisition system.}
	\label{fig:resultfail}
\end{figure*}

As a qualitative demonstration, in section \ref{subsec:realimg} we present some examples of layout reconstruction from real images, taken by the authors. We observe that the performance varies from environment to environment. We are able to recover the layout of all the environment and, in one of the cases, we achieve a good 3D scaled reconstruction of the layout.
Other examples with images from our proposed dataset are shown in Fig. \ref{fig:results}. We show examples of layouts where our method obtains good results in layout reconstruction and scale recovery. In the figure \ref{fig:resultfail} we show cases where the scale recovery of the environment fails. Analysing those cases, we found different possible sources of error. 
One of these sources is the information provided by the network, which may be noisy.
Other source of error is occlusion management in Atlanta environments. We found out that an incorrect definition of the 3D corners of the room when dealing with an occlusion lead the final adjustment to under-estimate the scale of the room. This seem logical since the closer the 3D points are to the acquisition reference system, the lower the reprojection error will be. 
A final source of error is the effective baseline of the non-central acquisition system. This problem is more evident in the real images, Fig \ref{fig:realExp}. With a smaller effective baseline, related with the radius of the non-central panorama, the accuracy to compute the scale of the room is reduced. 

\section{Conclusions}
\label{sec:conclusion}

In this paper we propose a new pipeline that completely solves the layout reconstruction problem from a single image (i.e. reconstructing Manhattan and Atlanta layouts with scale managing occlusions). We have presented the first application of non-central panoramas able to improve state-of-the-art methods for layout reconstruction. We use a neural network to obtain the boundaries of the structural lines of an indoor environment from a single non-central panorama, trained in a new dataset of non-central panoramas. Besides, we present two new geometrical solutions that process the output of the network to recover the 3D layout corners. From our full pipeline, we retrieve the 3D scaled layout, without extra measurements, and handle occluded walls in Manhattan and Atlanta environments. The experiments presented show that our method works in synthetic as well as real images.

\section*{ACKNOWLEDGMENT}

This work was supported by RTI2018-096903-B-100 (AEI/ FEDER, UE).

\bibliographystyle{unsrtnat}

\bibliography{biblo}

\end{document}